\documentclass{article}
\PassOptionsToPackage{numbers, compress}{natbib}
\usepackage[preprint]{neurips_2025}
\usepackage[utf8]{inputenc} %
\usepackage[T1]{fontenc}    %
\usepackage{url}            %
\usepackage{booktabs}       %
\usepackage{amsfonts}       %
\usepackage{nicefrac}       %
\usepackage{microtype}      %
\usepackage{graphicx} \graphicspath{{figures/}}
\usepackage{amsmath,amssymb,mathbbol}
\usepackage{bm}
\usepackage{tabularx,colortbl,multirow,array,makecell}
\usepackage{algorithmic}
\usepackage[linesnumbered,ruled,vlined]{algorithm2e}
\usepackage{acronym}
\usepackage{enumitem}
\usepackage[dvipsnames]{xcolor}
\usepackage[pagebackref,breaklinks,colorlinks,citecolor=gray]{hyperref}
\usepackage{xspace}
\usepackage[skip=3pt,font=small]{subcaption}
\usepackage[skip=3pt,font=small]{caption}
\usepackage[capitalise]{cleveref}
\usepackage[misc]{ifsym}
\usepackage{pifont}
\usepackage{natbib}
\setcitestyle{authoryear,square,citesep={;},aysep={,},yysep={;}}

\makeatletter
\DeclareRobustCommand\onedot{\futurelet\@let@token\@onedot}
\def\@onedot{\ifx\@let@token.\else.\null\fi\xspace}
\def\eg{\emph{e.g}\onedot} 

\def\ie{\emph{i.e}\onedot}

\makeatother

\crefname{algorithm}{Alg.}{Algs.}
\Crefname{algocf}{Algorithm}{Algorithms}
\crefname{section}{Sec.}{Secs.}
\Crefname{section}{Section}{Sections}
\crefname{table}{Tab.}{Tabs.}
\Crefname{table}{Table}{Tables}
\crefname{figure}{Fig.}{Fig.}
\Crefname{figure}{Figure}{Figure}

\makeatletter
\renewcommand{\paragraph}{%
  \@startsection{paragraph}{4}%
  {\z@}{0ex \@plus 0ex \@minus 0ex}{-1em}%
  {\hskip\parindent\normalfont\normalsize\bfseries}%
}
\makeatother

\newcommand{\norm}[1]{\left\Vert #1 \right\Vert}

\definecolor{wcolor}{RGB}{251,76,31}

\acrodef{hsi}[HSI]{Human-Scene Interaction}
\acrodef{hoi}[HOI]{Human-Object Interaction}
\acrodef{nerf}[NeRF]{Neural Radiance Field}
\acrodef{mlp}[MLP]{multilayer perceptron}
\acrodef{cse}[CSE]{Continuous Surface Embedding}

\newcommand{\append}{\textit{Appendix}\xspace}
\newcommand{\dataset}{\textbf{\textit{DogMo}}\xspace}
\newcommand{\ndogs}{$10$\xspace}
\newcommand{\ncameras}{$5$\xspace}
\newcommand{\nactions}{$11$\xspace}
\newcommand{\nframes}{$1M$\xspace}
\newcommand{\nminutes}{$220$\xspace}
\newcommand{\nmotions}{$1.2k$\xspace}

\title{\dataset: A Large-Scale Multi-View RGB-D Dataset for 4D Canine Motion Recovery}

\author{%
    \bf Zan Wang$^{1\,*}$, Siyu Chen$^{1\,*}$, Luya Mo$^{1\,*}$, Xinfeng Gao$^{1\,*}$\\
    \bf Yuxin Shen$^{1}$, Lebin Ding$^{1}$, Wei Liang$^{1,4\,\textrm{\dag}}$\\
    $\,*$ Equal contributors \quad $\,\textrm{\dag}$ Corresponding authors\\
    $^1$ School of Computer Science \& Technology, Beijing Institute of Technology\\
    $^2$ Yangtze Delta Region Academy of Beijing Institute of Technology, Jiaxing\\
    \vspace{-3pt}\\
    \url{https://pie-lab.cn/DogMo/}
}

\begin{document}
\maketitle

\begin{figure}[ht!]
    \centering
    \includegraphics[width=\linewidth]{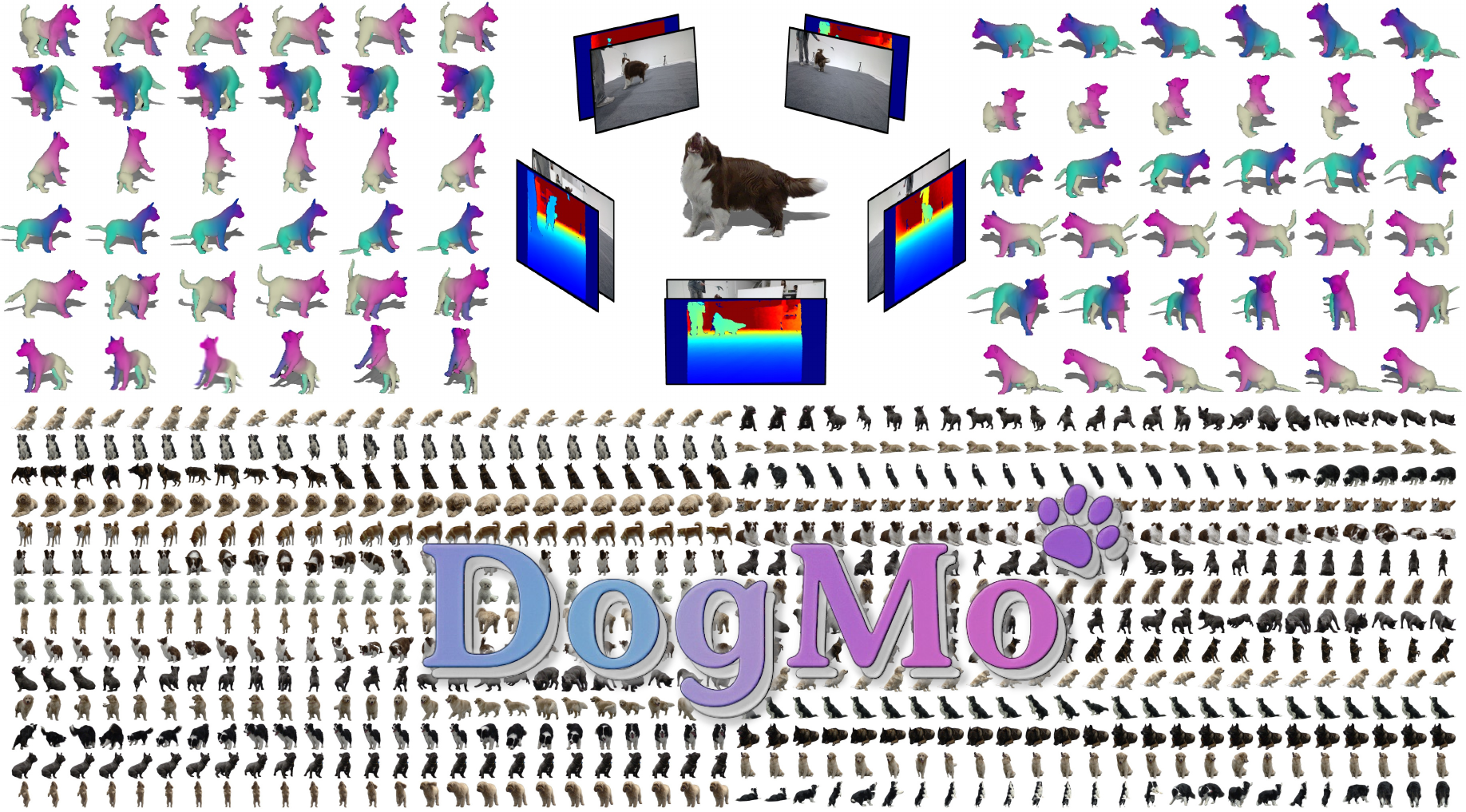}
    \caption{We present \dataset, a large-scale multi-view RGB-D video dataset capturing diverse canine movements using \ncameras synchronized low-cost cameras. \dataset contains \nmotions motion sequences from \ndogs unique dog instances, spanning \nactions action types and totaling over \nminutes minutes of recordings.}
    \label{fig:teaser}
\end{figure}

\begin{abstract}
We present \dataset, a large-scale multi-view RGB-D video dataset capturing diverse canine movements for the task of motion recovery from images. \dataset comprises \nmotions motion sequences collected from \ndogs unique dogs, offering rich variation in both motion and breed. It addresses key limitations of existing dog motion datasets, including the lack of multi-view and real 3D data, as well as limited scale and diversity. Leveraging \dataset, we establish four motion recovery benchmark settings that support systematic evaluation across monocular and multi-view, RGB and RGB-D inputs.
To facilitate accurate motion recovery, we further introduce a three-stage, instance-specific optimization pipeline that fits the SMAL model to the motion sequences. Our method progressively refines body shape and pose through coarse alignment, dense correspondence supervision, and temporal regularization. Our dataset and method provide a principled foundation for advancing research in dog motion recovery and open up new directions at the intersection of computer vision, computer graphics, and animal behavior modeling.
\end{abstract}

\section{Introduction}

Over the past decade, human body recovery from images and videos has made remarkable progress, driven by the release of numerous large-scale public datasets~\citep{mahmood2019amass,black2023bedlam}, standardized benchmarks~\citep{ionescu2013human3,von2018recovering}, and advanced algorithmic frameworks~\citep{bogo2016keep,kanazawa2018end,li2022cliff,shin2024wham}. As interest in animal welfare and the pet industry continues to grow, efforts have emerged to extend human motion recovery methods to animals~\citep{zuffi20173d,biggs2020left,ruegg2022barc,sabathier2024animal} for behavior analysis and virtual asset creation. However, progress in this area remains limited, primarily due to the lack of comprehensive animal datasets, which constrains advancements in algorithm design and evaluation.

Acquiring animal motion data is inherently inefficient and costly due to the lack of a tailored toolchain and the fact that animals, unlike humans, cannot follow instructions consistently during capture. Consequently, publicly available animal datasets are scarce and suffer from two major issues:
\begin{itemize}[leftmargin=*,nolistsep,noitemsep]
\item \textbf{Lack of multi-view and real 3D.} Most datasets~\citep{biggs2019creatures,yu2021benchmark,yang2022apt} consist of monocular RGB videos, lacking both multi-view recordings and accurate 3D information, which limits the study of multi-view settings and the recovery of real 3D shapes.
\item \textbf{Limited scale and diversity.} Existing datasets~\citep{kearney2020rgbd,luo2022artemis,shooter2024digidogs,shooter2024sydog} typically contain less than one hour of footage, with minimal variation in motion types and animal instances.
\end{itemize}

In this work, we focus on motion recovery for dogs, one of the most common companion animals in human life. To address the aforementioned issues, we introduce \dataset, a large-scale multi-view RGB-D video dataset capturing dog movements, collected using \ncameras low-cost RGB-D cameras that are carefully synchronized and calibrated, as illustrated in \cref{fig:teaser}. These cameras are evenly distributed around a capture zone to record dog movements from multiple viewpoints. Within this zone, dogs are encouraged to perform natural behaviors such as running, jumping, and interacting with humans.
In total, \dataset collects \nmotions motion sequences from \ndogs unique dog instances, with a total duration exceeding \nminutes minutes. Built upon this dataset, we establish four canine motion recovery settings, \ie, recovering 4D dog motions from (1) monocular RGB videos, (2) multi-view RGB videos, (3) monocular RGB-D videos, and (4) multi-view RGB-D videos.

Despite significant progress in human motion recovery, estimating dog motion from visual observations presents unique challenges. First, the wide variation in body size across breeds limits the model's generalization capability. Yet prior works~\citep{zuffi20173d,biggs2020left,yang2023ppr} primarily address monocular shape estimation, which inherently suffers from ambiguities in recovering absolute body shape. Second, dogs' fur-covered and often textureless appearance makes body parts visually indistinct, hindering accurate localization. Approaches such as \citet{ruegg2022barc,sabathier2024animal} depend on pre-trained keypoint detectors or part segmentation models, but they overlook the potential of temporal motion priors to regularize parts and joints across time.

To address these challenges, we first extend D-SMAL~\citep{zuffi20173d,ruegg2022barc} with an additional scaling factor to better deal with the broad variation in dog body shapes. Building on this extension, we propose a three-stage instance-specific optimization pipeline that fits the model parameters to input images in a coarse-to-fine manner. The first stage performs a lightweight coarse optimization to initialize the global translation, orientation, and body shape, substantially reducing the optimization space and improving overall efficiency. In the second stage, following \citet{sabathier2024animal}, we leverage \ac{cse}~\citep{neverova2020continuous} for dense correspondence supervision, augmented with regularization terms to address the unique leg-crossing issue and to constrain shape and pose variations. Finally, the third stage refines the entire motion sequence with a temporal loss that enforces smoothness and mitigates local optima arising from noisy keypoint estimates in individual frames.

We summarize the contributions as: 
(1) We introduce \dataset, a large-scale multi-view RGB-D dataset for 4D dog motion recovery, addressing key limitations of existing datasets, including the lack of multi-view coverage and real 3D information, and sufficient data scale and diversity.
(2) Based on \dataset, we establish four benchmark settings spanning monocular and multi-view, RGB and RGB-D inputs, aiming to facilitate further research in animal motion recovery and related downstream applications.
(3) We propose a three-stage instance-specific optimization pipeline that progressively refines SMAL parameter estimation from coarse to fine.

\section{Related Work}

\paragraph{Model-Free 3D Animal Reconstruction}
Model-free 3D animal reconstruction from images or videos either employs instance-specific 3D shape optimization~\citep{cashman2012shape,yang2021lasr,wu2022casa,yang2023ppr} or trains neural networks to directly regress 3D structures without requiring 3D supervision~\citep{wu2023magicpony,liu2023lepard,wu2023dove,yao2023artic3d,aygun2024saor,li2024learning}. Both approaches avoid reliance on \textit{parametric models} and primarily leverage differentiable rendering to enforce 2D supervision, such as silhouettes, keypoints, and semantic correspondences. While the former is typically tailored to individual instances, the latter aims to generalize across diverse articulated animal species by learning from large-scale internet image collections.
More recent efforts have shifted toward building \textit{neural implicit representations} for animals~\citep{luo2022artemis,yang2022banmo,sinha2023common,tan2023distilling,tu2025dreamo,yang2025agent}, or distilling 3D reconstructions from 2D generative models~\citep{ho2020denoising,rombach2022high} by extending Score Distillation Sampling (SDS)~\citep{poole2023dreamfusion} to image-conditioned multi-view generation models~\citep{liu2023zero,qian2024magic123,shi2024mvdream}.
Our focus is on model-based 4D dog motion recovery from a wide range of visual inputs, spanning monocular and multi-view, RGB and RGB-D inputs.

\paragraph{Model-Based Animal Pose and Shape Estimation}
Model-based animal pose estimation from 2D observations, such as 2D~\citep{cao2019cross,yu2021benchmark,yang2022apt,zeng2024towards,ye2024superanimal} and 3D~\citep{kearney2020rgbd,bala2020automated,muramatsu2022improving,sosa2023horse,shooter2024digidogs} keypoint detection, plays a crucial role in animal behavior analysis.
Leveraging parametric animal models like SMAL~\citep{zuffi20173d} and ABM~\citep{badger20203d}, recent approaches aim to directly estimate pose and shape parameters from images, as these models provide strong priors over animal body shape and articulation. These methods typically either optimize~\citep{zuffi20173d,wang2021birds,li2021hsmal,sabathier2024animal} or regress~\citep{biggs2020left,li2021coarse,ruegg2022barc,ruegg2023bite,li2024dessie} the parameters from observations, often relying on 2D re-projection losses due to the absence of ground-truth 3D annotations. 
Some studies~\citep{zuffi2019three,lyu2024animer,shooter2024digidogs} instead train models to directly regress 3D pose and shape using synthetic data or motion capture datasets, though these approaches are often limited in data diversity and scale.
We propose a unified three-stage pipeline for progressively fitting the SMAL model to video sequences, demonstrating its robust performance across various visual inputs.

\paragraph{Animal-Related Datasets}
Early datasets for animal analysis mainly comprise 2D images or videos annotated with 2D labels, enabling tasks such as classification~\citep{khosla2011novel}, detection~\citep{liu2023lote}, segmentation~\citep{mu2020learning,deane2024rgbt}, 2D/3D pose estimation~\citep{biggs2019creatures,cao2019cross,biggs2020left,yu2021benchmark,joska2021acinoset,yang2022apt,yao2023openmonkeychallenge}, and behavior understanding~\citep{li2020atrw,ng2022animal,ma2023chimpact}.
Compared to these datasets, some works leverage simulators to synthesize animal images~\citep{zuffi2019three,shooter2024digidogs,shooter2024sydog,lyu2024animer}, which provide access to 3D pose. Other efforts rely on expensive capture devices, such as motion capture systems or 3D scanners, to acquire high-quality 3D data of horses~\citep{li2024poses,zuffi2024varen}, birds~\citep{naik20233d}, and rats~\citep{marshall2021pair}.
\citet{kearney2020rgbd} collect RGB-D data of dogs using a Kinect V2 camera, supplemented with MoCap to obtain accurate 3D poses. \citet{xu2023animal3d} generate pseudo ground-truth 3D annotations by manually fitting the SMAL model to 2D images. Rodent3D~\citep{patel2023animal} provides multi-view, multi-modal video recordings of mice to facilitate pose tracking and behavioral analysis. The emergence of \ac{nerf} has inspired datasets~\citep{luo2022artemis,sinha2023common} that leverage multi-view imaging or video to reconstruct detailed 3D neural animals.
This work introduces a new large-scale multi-view RGB-D video dataset capturing diverse canine movements, aimed at facilitating research in dog motion recovery and related fields.

\section{\dataset Dataset}

\subsection{Data Capture}

\begin{figure}[ht!]
    \centering
    \begin{subfigure}{0.5\linewidth}
        \centering
        \includegraphics[width=\linewidth]{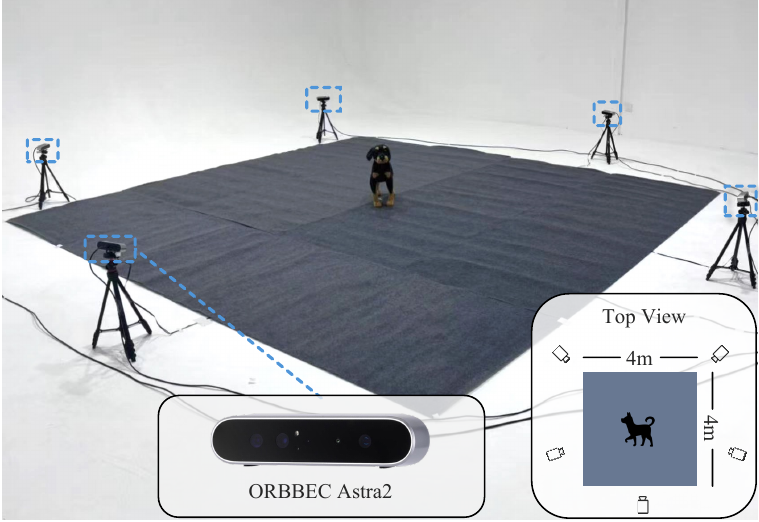}
        \caption{Capture System}
        \label{fig:device}
    \end{subfigure}\hfill%
    \begin{subfigure}{0.5\linewidth}
        \centering
        \includegraphics[width=\linewidth]{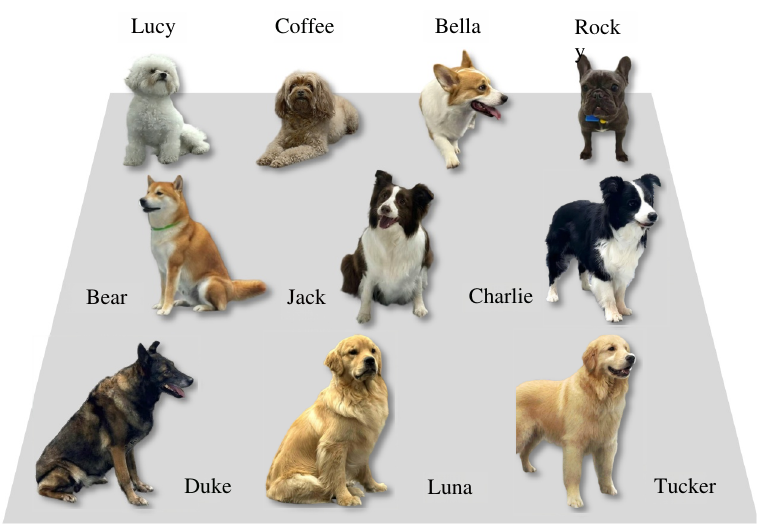}
        \caption{Dogs}
        \label{fig:dogs}
    \end{subfigure}
    \caption{\textbf{Motion Capture.} We employ \ncameras synchronized low-cost RGB-D cameras to record \ndogs individual dogs.}
    \label{fig:capture}
\end{figure}

\paragraph{Device Setup}
We collect dog motion data using a low-cost capture system equipped with \ncameras ORBBEC Astra 2\footnote{ORBBEC Astra 2: \url{https://www.orbbec.com/products/structured-light-camera/astra-2/}} RGB-D cameras. The cameras are evenly distributed along a circular area with a radius of approximately $2.5$ meters, all oriented toward the center of the capture zone. Each camera is positioned at a height of about $0.45$ meters, with a random pitch angle within $\pm 10$ degrees. \cref{fig:capture} exhibits the overview of the capture setup.

For camera calibration, we leverage the OpenCV library~\citep{bradski2000opencv} to manually calibrate each pair of adjacent cameras, thus deriving the relative poses of any two cameras. To synchronize all cameras, we employ the Sync Hub Dev\footnote{Sync Solutions: \url{https://www.orbbec.com/products/camera-accessories/sync-solutions/}} device provided by ORBBEC.

\paragraph{Capture Procedure} During data collection, a dog trainer guides the dog to perform specific actions within the capture zone by issuing semantic commands, such as running, jumping, or playing with humans. We ensure that the dog remains within the capture area so that at least three cameras can simultaneously capture the dog's full body. Since the trainer needs to actively guide the dog, occasional occlusions of the dog by the trainer are unavoidable from certain viewpoints.

\subsection{Dataset Statistics}
\begin{figure}[t!]
    \centering
    \includegraphics[width=\linewidth]{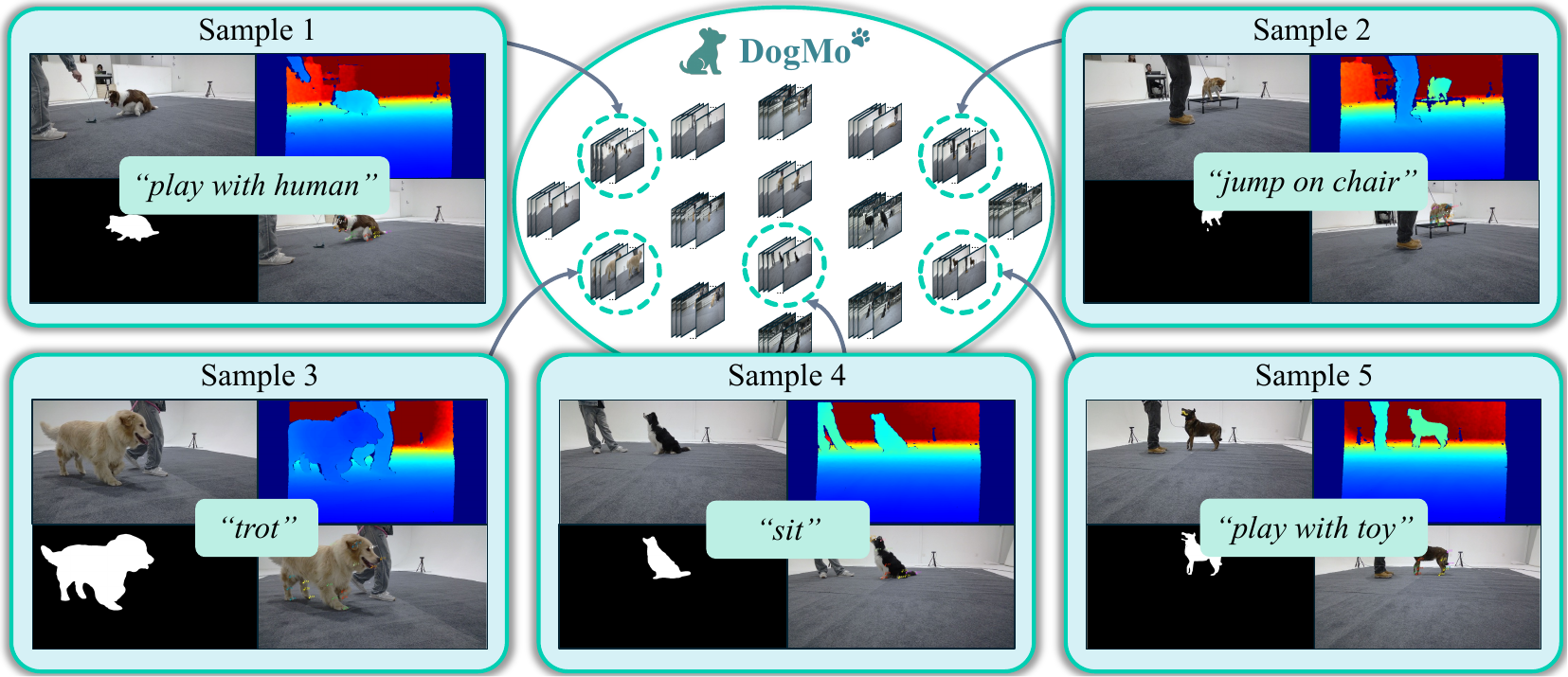}
    \caption{\textbf{Overview of \dataset.} \dataset provides five-view RGB-D image sequences for each motion recording, along with corresponding instance masks, keypoints, and semantic (\ie, action label) annotations.}
    \label{fig:dataset}
\end{figure}

In \dataset, each camera records synchronized RGB and depth streams, with RGB images at a resolution of $1920 \times 1080$ and depth images at $1600 \times 1200$. The RGB and depth frames are hardware-aligned within the camera. The system operates at $15$ FPS.
\dataset contains \nmotions motion clips collected from \ndogs unique dog instances, as displayed in \cref{fig:dogs}. Each motion clip lasts between $0.87$ and $19.07$ seconds, covering \nactions action types ranging from running to playing with humans. In total, the dataset provides over \nminutes minutes of motion footage, amounting to more than \nframes frames across all camera views.
To further support research, we annotate each frame with binary masks and keypoints of the dog, and assign a semantic label to each motion. We provide an overview of \dataset in \cref{fig:dataset}, and a comparison with existing related datasets is summarized in \cref{tab:dataset_comparison}.

\begin{table}[ht!]
    \centering
    \small
    \setlength{\tabcolsep}{3pt}
    \caption{\textbf{Comparison between \dataset and existing related datasets.} The notation $^{*}$ indicates that nearly all dog motions in the CoP3D dataset are in a sprawling posture throughout the entire video.}
    \label{tab:dataset_comparison}
    \resizebox{\linewidth}{!}{%
        \begin{tabular}{lcccccccc}%
            \toprule
            Dataset  & Modality & FPS & Resolution & \#Views & \#Frames & \#Subjects & \#Actions & Sync. / Real \\
            \midrule
            CoP3D~\citep{sinha2023common}        & RGB   & -    & $720\times1280$  & $1$ & $600k$  & $4,200$ & $1^{*}$ & Real  \\
            Sydog-video~\citep{shooter2024sydog} & RGB   & $25$ & $1024\times1024$ & $1$ & $87.5k$ & $5$     & $5$     & Sync. \\
            Digidogs~\citep{shooter2024digidogs} & RGB-D  & -    & $1280\times720$  & $1$ & $27.9k$ & $8$     & $12$    & Sync. \\
            RGBD-Dogs~\citep{kearney2020rgbd}    & RGB-D  & $6$  & $2560\times1440$ & $6$ & $1.9k$  & $7$     & $5$     & Real  \\
            \dataset (Ours)                      & RGB-D  & $15$ & $1920\times1080$ & \ncameras & \nframes & \ndogs & \nactions & Real \\
            \bottomrule
        \end{tabular}%
    }%
\end{table}

\section{Method}

\subsection{Problem Formulation}

Given a video sequence $\{\textbf{I}_t\}_{t=1}^T$, our goal is to reconstruct the 3D posed mesh of a canine subject, parameterized by D-SMAL~\citep{zuffi20173d,ruegg2023bite} parameters $\{\textbf{S}_t\}_{t=1}^T$. Each frame $\textbf{I}_t$ can be a single-view RGB, multi-view RGB, single-view RGB-D, or multi-view RGB-D input, where $T$ denotes the number of frames.
For simplicity, we denote the RGB and depth images as $C_t$ and $D_t$, respectively.
The D-SMAL parameters at each frame are represented as a $5$-tuple $\textbf{S}_t = (\beta, \theta_t, \gamma_t, \phi_t, s)$, where $\beta \in \mathbb{R}^{30}$ represents the shape parameters, $\theta_t \in \mathbb{R}^{6N}$ denotes the $N$ joint rotations in $6$D representation, $\gamma_t \in \mathbb{R}^{3}$ is the global translation, and $\phi_t \in \mathbb{R}^{6}$ is the global orientation. To accommodate the broader range of dog sizes, we introduce $s$ as an additional scaling parameter for resizing the body mesh, extending the original D-SMAL formulation.
Given these parameters, the D-SMAL model finally outputs a posed 3D mesh, defined as
\begin{equation}
\mathcal{V}, \mathcal{F} = \mathcal{M}(\beta, \theta, \phi) \times s + \gamma,
\end{equation}
where $\mathcal{V} \in \mathbb{R}^{3889 \times 3}$ are the mesh vertices, $\mathcal{F} \in \mathbb{N}^{7774 \times 3}$ is a list of triangular faces, and $\mathcal{M}(\cdot)$ is original skinning function of D-SMAL.

\subsection{Canine Motion Recovery}

\begin{figure}[t!]
    \centering
    \includegraphics[width=\linewidth]{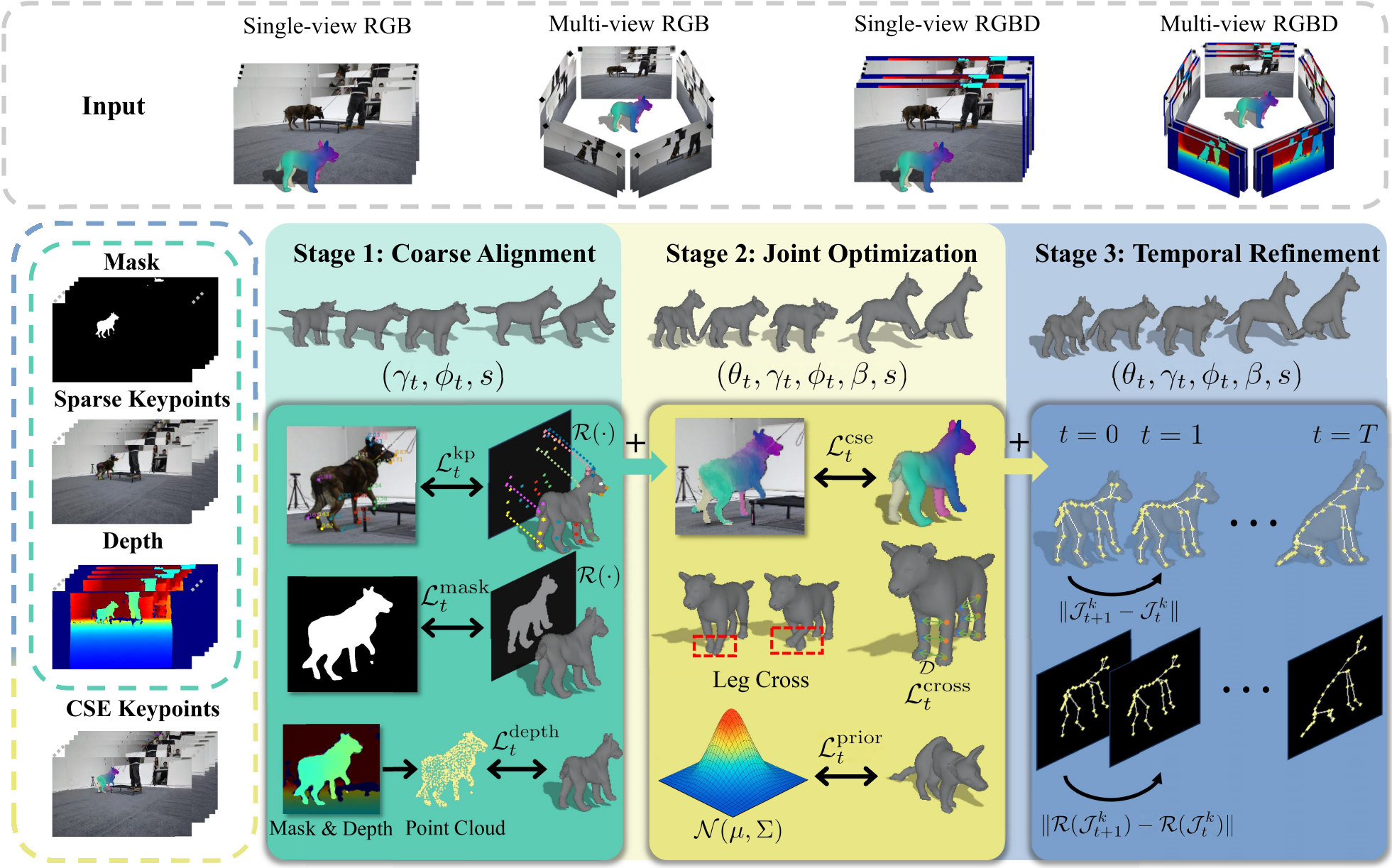}
    \caption{\textbf{Optimization pipeline.} Our optimization pipeline includes three stages, progressively refining the dog's body shape and pose through coarse alignment, dense correspondence supervision, and temporal regularization.}
    \label{fig:model}
\end{figure}

To encourage temporal smoothness of the results, we follow \citet{sabathier2024animal} and model the time-varying tuple $(\theta_t, \gamma_t, \phi_t)$ as the output of an \ac{mlp} network conditioned on a smooth timestep embedding $\tau(t)$, instead of directly optimizing the D-SMAL parameters. Here, 
\begin{equation}
\theta_t, \gamma_t, \phi_t = \mathrm{MLP}_{\psi}(\tau(t)).
\end{equation}
In contrast, the body shape parameters $\beta$ and the scaling factor $s$ are treated as time-invariant and are directly optimized.
During the optimization, we adopt a three-stage optimization pipeline that progressively recovers the \ac{mlp} parameters $\psi$, body shape parameter $\beta$, and the scaling factor $s$, from coarse to fine, as detailed below. \cref{fig:model} presents the overview of the optimization pipeline.

\subsubsection{Stage 1: Coarse Alignment.}
\label{sec:stage_one}

In the first stage, we initialize the parameters $(\gamma_t, \phi_t, s)$ by coarsely aligning the projected 3D model with the observed data. This alignment is guided by three frame-specific loss terms defined below.

\paragraph{Chamfer Mask Loss}
The Chamfer Mask Loss encourages alignment between the projected D-SMAL mesh and the observed dog silhouette, defined as:
\begin{equation}
\mathcal{L}_t^{\mathrm{mask}} = \mathrm{CD}\left(\{u \mid u \in C_t \odot M_t\}, \{\mathcal{R}(X) \mid X \in (\mathcal{V}_t, \mathcal{F}_t)\}\right).
\end{equation}
Here, $u$ denotes the 2D pixels occupied by the dog, derived from the segmentation mask $M_t$; and $M_t$ is predicted by the off-the-shelf Grounded SAM 2~\citep{ren2024grounded} using the text prompt ``dog.''
$\mathcal{R}(\cdot)$ denotes the camera projection operation, which projects 3D points onto the 2D image plane using a fixed projection matrix shared across all frames under the assumption of a static camera setup.
$X$ represents 3D points sampled from the posed mesh $(\mathcal{V}_t, \mathcal{F}_t)$. To better capture leg motion, we increase the sampling density in regions corresponding to legs, improving their silhouette alignment.

\paragraph{Sparse Keypoint Loss}
To further enforce alignment between the keypoints of the 3D model and observed 2D keypoints, we apply the Sparse Keypoint Loss defined as:
\begin{equation}
\mathcal{L}^{\mathrm{kp}}_t = \norm{h(C_t) - \mathcal{R}(\mathcal{J}_t)},
\end{equation}
where $h(C_t)$ denotes the 2D keypoints detected from the image $C_t$ using the sparse dog keypoint detector from BARC~\citep{ruegg2022barc}, and $\mathcal{R}(\mathcal{J}_t)$ represents the projection of the corresponding 3D keypoints $\mathcal{J}_t$ from the estimated D-SMAL mesh.

\paragraph{Chamfer Depth Loss}
To exploit per-pixel depth supervision when depth input is available, we incorporate a Chamfer Depth Loss defined as:
\begin{equation}
\mathcal{L}_t^{\mathrm{depth}} = \mathrm{CD}(\{v \mid v \in \mathcal{R}^{-1}( M_t \odot D_t)\}, \{X \mid X \in (\mathcal{V}_t, \mathcal{F}_t)\}.
\end{equation}
Here, $\mathcal{R}^{-1}$ denotes the inverse camera projection operation, which uses the depth map and the camera projection matrix to lift the masked depth image pixels into 3D space. This loss measures the Chamfer Distance between the resulting point cloud and the sampled points on the estimated dog mesh. We use this term in settings that include depth input.

\subsubsection{Stage 2: Joint Optimization.}

In the second stage, we optimize all parameters through a more comprehensive, frame-specific loss function, which includes three additional loss terms beyond the previously introduced terms. 

\paragraph{CSE Keypoint Loss}
Following \citet{sabathier2024animal}, we incorporate a CSE Keypoint Loss to enforce denser correspondence supervision. This loss is defined as:
\begin{equation}
\mathcal{L}^{\mathrm{cse}}_t = \sum_{u \in M_t \odot C_t} \left\| u - \mathcal{R}(\mathrm{CSE}(\mathcal{V}, u) \right\|,
\end{equation}
which penalizes the distance between each foreground pixel $u$ in the dog mask and the 2D projection of its corresponding 3D point on the D-SMAL mesh. The correspondence is provided by the image-to-CSE predictor $\mathrm{CSE}(\cdot)$~\citep{neverova2020continuous}.

\paragraph{Leg Cross Loss}
To prevent unnatural intersections between the left and right legs of the posed D-SMAL model, we introduce a Leg Cross Loss defined as:
\begin{equation}
\mathcal{L}_{t}^{\mathrm{cross}}=\sum_{(J^l, J^r) \in \mathcal{J}_t^{\mathrm{foot}}} \mathbb{1}_{\geq \delta} \cdot \exp(-\|(J^l - J^r)\|,
\end{equation}
which penalizes overly close joint pairs from opposite legs to ensure physically plausible and natural leg motion. Here, $\mathcal{J}_t^{foot}$ denotes a predefined set of joint pairs corresponding to the left and right legs, and $\mathbb{1}_{\geq \delta}$ is an indicator function that outputs $1$ when $\exp(-\|(J^l - J^r)\|$ exceeds threshold $\delta$, and $0$ otherwise.

\paragraph{Shape \& Pose Prior Loss}
To impose structural and kinematic constraints on the deformed mesh, we follow \citet{ruegg2022barc, zuffi20173d} and apply a prior loss defined as:
\begin{equation}
\mathcal{L}^{\mathrm{prior}} = \mathcal{L}^{\mathrm{shape}} + \sum_{t=1}^{T}\mathcal{L}_t^{\mathrm{pose}}.
\end{equation}
The shape prior term $\mathcal{L}^{\mathrm{shape}}$ encourages the D-SMAL shape coefficients $\beta$ to match a prior distribution.
The pose prior term $\mathcal{L}_t^{\mathrm{pose}}$ constrains the estimated D-SMAL pose parameters $\theta_t$ to remain within plausible limits by applying a Gaussian prior penalty. Details are presented in \append.

\subsubsection{Stage 3: Temporal Refinement.}

While the previous loss terms are computed independently for each frame, they do not enforce temporal consistency across the sequence. The third stage incorporates a Temporal Smoothness Loss, in addition to the above comprehensive frame-wise losses, to further refine the motion sequence:
\begin{equation}
\mathcal{L}^{\mathrm{temp}} = \sum_{t=1}^{T-1} \sum_{k=1}^{|\mathcal{J}|} \left(\|\mathcal{J}_{t+1}^k - \mathcal{J}_{t}^k\| + \|\mathcal{R}(\mathcal{J}_{t+1}^k) - \mathcal{R}(\mathcal{J}_{t}^k)\|\right),
\end{equation}
which penalizes abrupt changes in both 3D joint positions and their 2D projections.

\subsection{Implementation Details}
\label{sec:implementation}
\paragraph{Method Usage Across Different Settings}
In the first and second stages, the loss is computed by averaging over mini-batches of frames sampled from video sequences. In the third stage, motion is further refined by optimizing over a sampled continuous sequence segment. Based on \dataset, we establish four evaluation setups for motion recovery. For setups involving multi-view inputs, we average the loss across all views. The key difference between setups with and without depth input lies in using Chamfer Depth Loss, which is applied only when depth data is available.

\paragraph{Hyperparameter Configurations} In our implementation, we employ two separate \acp{mlp} to map the time-varying global translation and orientation, as well as body pose, \ie, $\mathrm{MLP}_{TR}$ and $\mathrm{MLP}_{\theta}$. $\mathrm{MLP}_{TR}$ consists of $3$ layers with $16$ hidden units per layer, while $\mathrm{MLP}_{\theta}$ has $3$ layers with $64$ hidden units per layer.
The number of optimization steps is the product of a specified multiplier and the frame number. For single-view sequences, we set the multipliers as $5$, $20$, and $5$ for the first, second, and third stages, respectively; for multi-view sequences, the corresponding multipliers are $10$, $25$, and $5$.
We perform optimization using the Adam optimizer. All experiments are implemented in PyTorch and run on a single NVIDIA GeForce RTX 4090 GPU. Refer to \append for more details.

\section{Experiments}
\label{sec:exp}

\subsection{Evaluation Setup}

\paragraph{Metrics}
We evaluate the methods using six metrics: (1) \textit{IoU}, the Intersection-over-Union between the rendered silhouette of the estimated SMAL model and the ground-truth dog segmentation; (2) \textit{IoU$_{\text{w5}}$}, the average IoU over the worst $5\%$ of frames in each sequence, reflecting the method's robustness; (3) \textit{F-score}~\citep{wang2018pixel2mesh}, which quantifies the distance between the estimated dog mesh and the inverse-projected dog point cloud from RGB-D images; (4) \textit{Pene$_\%$}, the proportion of mesh vertices that penetrate the floor over a sequence; (5) \textit{Jitter} (m), measuring temporal smoothness; and (6) \textit{Foot Skating (FS)} (m), which evaluates the naturalness of foot contact by capturing sliding artifacts.
\textbf{Given that the \textit{F-score} measures 3D reconstruction accuracy, we adopt it as the primary evaluation metric.}
The computation details of these metrics are presented in \append.

\paragraph{Baselines}
To the best of our knowledge, there are no open-source baselines available for dog motion recovery from multi-view or RGB-D videos. We adopt SMALify~\citep{biggs2020left}, a learning-based method that estimates dog pose from monocular RGB images, as the baseline for comparison on the single-view RGB setting. We use the officially released model and checkpoint to evaluate on our dataset.
We also compare against AnimalAvatar~\citep{sabathier2024animal}, an optimization-based method originally designed for recovering a dog model from single-view RGB video sequences, which we further adapt to handle RGB-D input. We use the default model configuration and hyperparameters.
In addition, we include two ablated variants of our pipeline, \textit{w/o stage 1} and \textit{w/o stage 3}, to assess the contribution of coarse alignment initialization and temporal refinement, respectively.

\subsection{Results}

\begin{figure}[t!]
    \centering
    \begin{subfigure}{\linewidth}
        \centering
        \includegraphics[width=\linewidth]{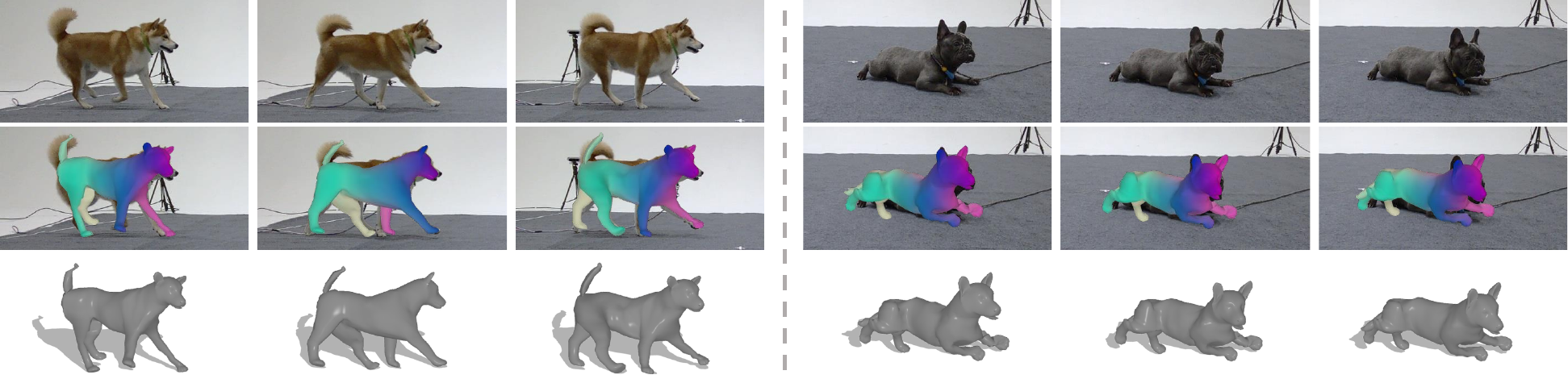}
        \caption{Optimization Results on Single-view RGB Setting}
        \label{fig:sg_rgb_results}
    \end{subfigure}\hfill%
    \vspace{3pt}
    \begin{subfigure}{\linewidth}
        \centering
        \includegraphics[width=\linewidth]{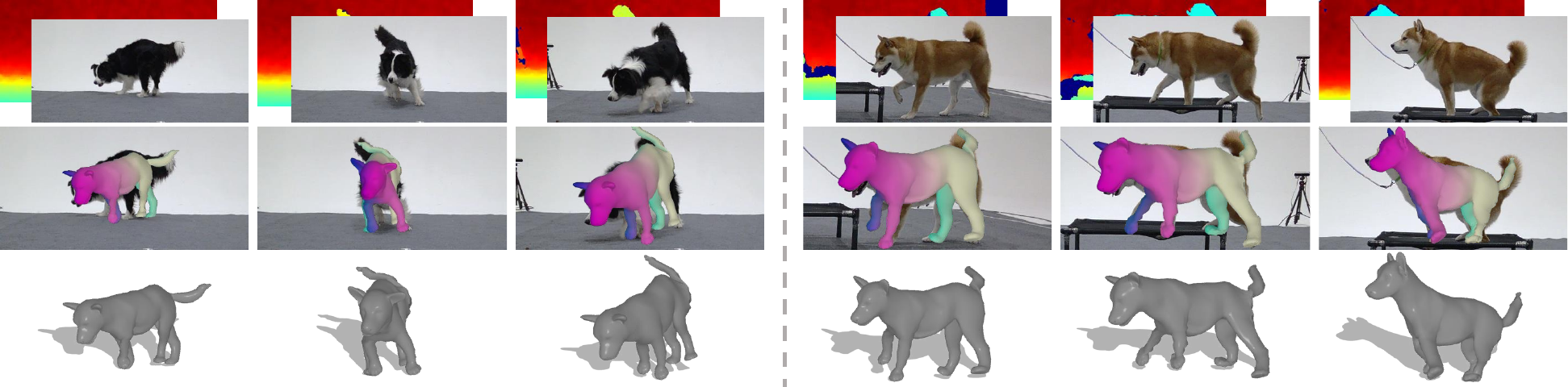}
        \caption{Optimization Results on Single-view RGB-D Setting}
        \label{fig:sg_rgbd_results}
    \end{subfigure}\hfill%
    \vspace{3pt}
    \begin{subfigure}{\linewidth}
        \centering
        \includegraphics[width=\linewidth]{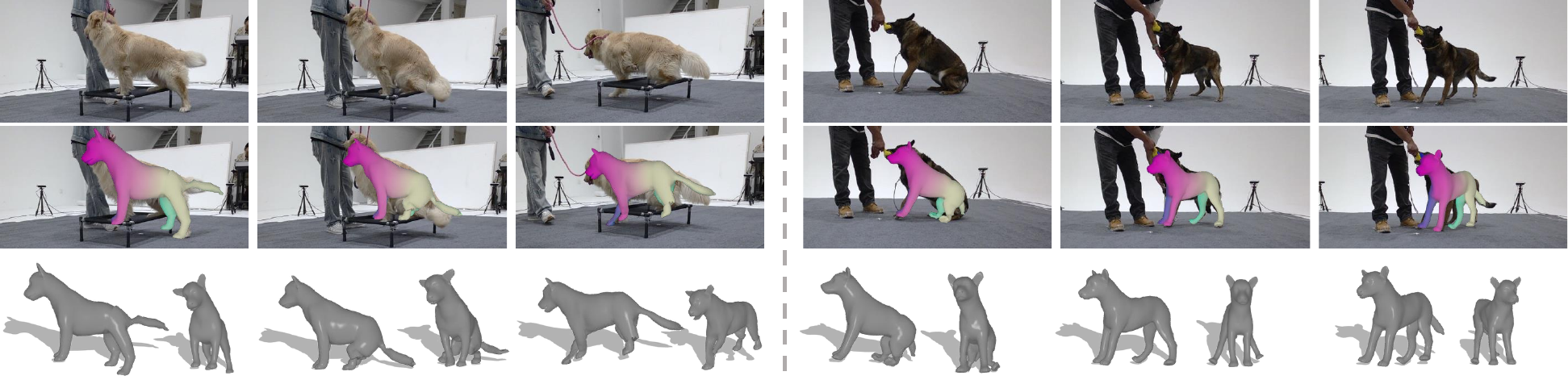}
        \caption{Optimization Results on Multi-view RGB Setting}
        \label{fig:ml_rgb_results}
    \end{subfigure}\hfill%
    \vspace{3pt}
    \begin{subfigure}{\linewidth}
        \centering
        \includegraphics[width=\linewidth]{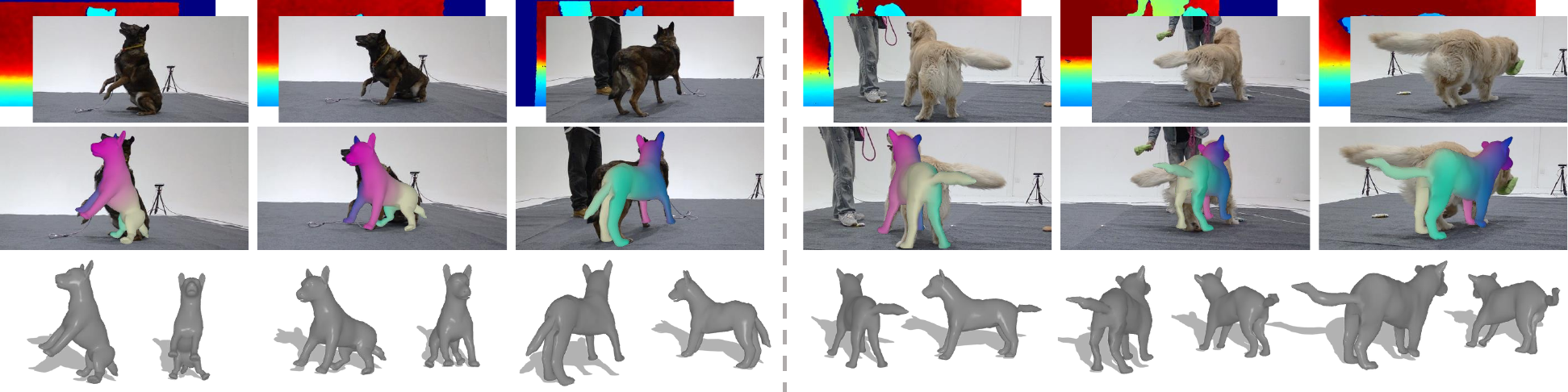}
        \caption{Optimization Results on Multi-view RGB-D Setting}
        \label{fig:ml_rgbd_results}
    \end{subfigure}\hfill%
    \caption{\textbf{Qualitative results.} Each benchmark setting shows two results, with three sampled frames per result from left to right. Three rows show the input image, \ac{cse} model projection, and recovered mesh, respectively. In multi-view settings, we show only one input view for clarity and visualize the mesh from two angles.}
    \label{fig:qual_results}
\end{figure}

\paragraph{Qualitative Results}
\cref{fig:qual_results} shows the qualitative results of our method across the four setups. Our method effectively recovers dog motion from various input types, ranging from single-view to multi-view and from RGB to RGB-D inputs.

\begin{figure}[t!]
    \centering
    \includegraphics[width=\linewidth]{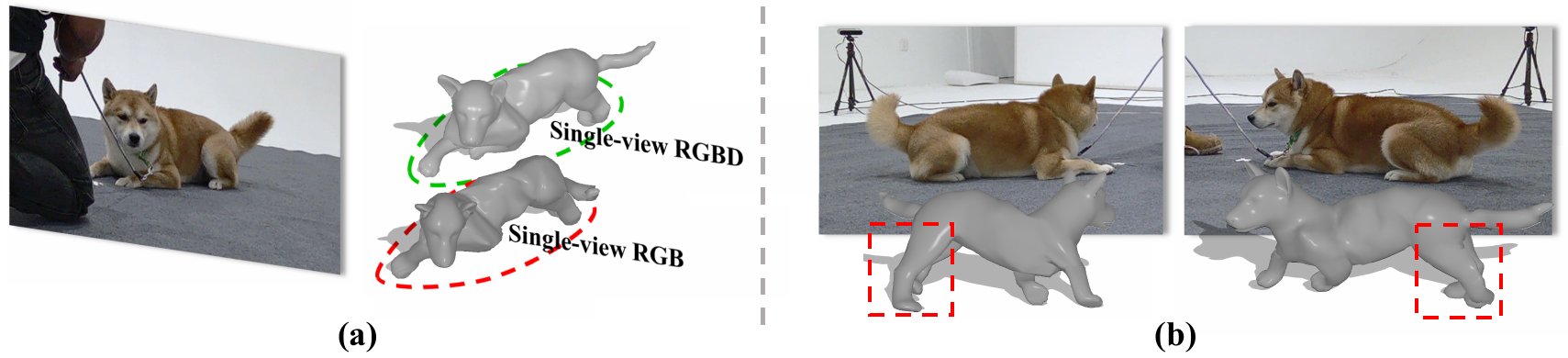}
    \caption{\textbf{Ambiguity in dog mesh recovery.} (a) Without real 3D information, the optimization method using only single-view RGB input fails to recover the dog's absolute position and size accurately. (b) The resulting mesh also exhibits noticeable artifacts when viewed from other perspectives.}
    \label{fig:issue}
\end{figure}

\paragraph{Quantitative Results}
We present the quantitative results in \cref{tab:quan_results}, from which we draw the following several key observations.
\begin{itemize}[leftmargin=*,nolistsep,noitemsep]
    \item Although the models under the single-view RGB setting achieve higher IoU scores, they consistently perform significantly worse in terms of F-score, Pene$_\%$, Jitter, and Foot Skating. This discrepancy suggests that single-view RGB lacks sufficient 3D cues to recover absolute body shape and position accurately in 3D space. In contrast, multi-view and depth inputs provide richer spatial information, enabling more precise reconstruction of 3D shape and position. These results are consistent with the qualitative visualizations in \cref{fig:issue}.
    \item Our method significantly outperforms SMALify~\citep{biggs2020left} and AnimalAvatar~\citep{sabathier2024animal} across IoU and F-score, demonstrating that these baselines struggle to accurately fit dog silhouettes and recover global 3D positions and body sizes. Although AnimalAvatar uses a similar optimization strategy, its reliance on the original D-SMAL model limits its ability to handle large variations in dog body sizes. Its one-stage, frame-independent optimization also lacks proper initialization and temporal constraints, often yielding suboptimal results. While AnimalAvatar yields better Pene$_\%$ and Foot Skating, this is primarily because it produces meshes that float above the floor. \textbf{Such non-contact meshes naturally result in zero penetration and foot skating, which spuriously improve these metrics.}
    \item The first-stage coarse alignment is crucial for motion recovery, particularly in the single-view RGB-D setting, as it provides favorable initialization and prevents suboptimal in the subsequent dense optimization. We also observe that varying the initial value of $s$ greatly affects the results of the model \textit{w/o stage 1}, especially for Pene$_\%$, Jitter, and FS; therefore, we omit these results in \cref{tab:quan_results}. Additional analysis is provided in \append. The third-stage temporal refinement further improves the final results in the single-view RGB setting, and its impact diminishes in multi-view or depth-rich scenarios, where the input provides sufficient information for accurate recovery.
\end{itemize}

\begin{table}[t!]
    \centering
    \small
    \setlength{\tabcolsep}{3pt}
    \caption{\textbf{Quantitative results.} \textbf{Bold} indicates the best result.}
    \label{tab:quan_results}
    \resizebox{\linewidth}{!}{%
        \begin{tabular}{c|cccccccccccc}%
            \toprule
            \multirow{2.4}{*}{Method} & \multicolumn{6}{c}{Single-view RGB} & \multicolumn{6}{c}{Single-view RGB-D} \\
            \cmidrule(lr){2-7} \cmidrule(lr){8-13} & IoU $\uparrow$ & $\text{IoU}_{\text{w5}}$ $\uparrow$ & F-score $\uparrow$ & Pene$_\%$ $\downarrow$ & Jitter $\downarrow$ & FS $\downarrow$ & IoU $\uparrow$ & $\text{IoU}_{\text{w5}}$ $\uparrow$ & F-score $\uparrow$ & Pene$_\%$ $\downarrow$ & Jitter $\downarrow$ & FS $\downarrow$ \\
            \midrule
            SMALify             & $0.6133$          & $0.4058$          & $0.2324$           & $0.4363$          & $0.0895$         & $0.0435$
                                & -                 & -                 & -                  & -                 & -                & -                 \\
            AnimalAvatar        & $0.6116$          & $0.5197$          & $0.0283$           & $\mathbf{0.2702}$ & $0.0298$         & $\mathbf{0.0144}$
                                & $0.6036$          & $0.4905$          & $0.4748$           & $0.2039$          & $0.0260$         & $0.0184$          \\
            w/o stage 1 (Ours)  & $0.6024$          & $0.4842$          & $0.3360$           & - & - & - 
                                & $0.6074$          & $0.4901$          & $0.6619$           & - & - & - \\
            w/o stage 3 (Ours)  & $0.6491$          & $0.5235$          & $0.3576$           & $0.3698$          & $0.0211$         & $0.0366$
                                & $\mathbf{0.6424}$ & $\mathbf{0.5235}$ & $0.7666$           & $0.1261$          & $0.0140$         & $0.0167$          \\
            Full (Ours)         & $\mathbf{0.6505}$ & $\mathbf{0.5247}$ & $\mathbf{0.3593}$  & $0.3653$          & $\mathbf{0.0206}$& $0.0361$
                                & $0.6414$          & $0.5228$          & $\mathbf{0.7673}$  & $\mathbf{0.1278}$ & $\mathbf{0.0137}$& $\mathbf{0.0166}$ \\
            \midrule
            \multirow{2.4}{*}{Method} & \multicolumn{6}{c}{Multi-view RGB} & \multicolumn{6}{c}{Multi-view RGB-D} \\
            \cmidrule(lr){2-7} \cmidrule(lr){8-13} & IoU $\uparrow$ & $\text{IoU}_{\text{w5}}$ $\uparrow$ & F-score $\uparrow$ & Pene$_\%$ $\downarrow$ & Jitter $\downarrow$ & FS $\downarrow$ & IoU $\uparrow$ & $\text{IoU}_{\text{w5}}$ $\uparrow$ & F-score $\uparrow$ & Pene$_\%$ $\downarrow$ & Jitter $\downarrow$ & FS $\downarrow$ \\
            \midrule
            w/o stage 1 (Ours)  & $0.6084$          & $0.4358$          & $0.9096$           & - & - & - 
                                & $0.6122$          & $0.4403$          & $0.9134$           & - & - & - \\
            w/o stage 3 (Ours)  & $\mathbf{0.6271}$ & $\mathbf{0.4496}$ & $\mathbf{0.9107}$  & $0.0700$          & $0.0124$         & $\mathbf{0.0188}$
                                & $\mathbf{0.6300}$ & $\mathbf{0.4557}$ & $\mathbf{0.9137}$  & $0.0699$          & $0.0123$         & $\mathbf{0.0188}$  \\
            Full (Ours)         & $0.6258$          & $0.4490$          & $\mathbf{0.9107}$  & $\mathbf{0.0688}$ & $\mathbf{0.0120}$& $\mathbf{0.0188}$
                                & $0.6286$          & $0.4549$          & $\mathbf{0.9137}$  & $\mathbf{0.0687}$ & $\mathbf{0.0119}$& $0.0187$           \\
            \bottomrule
        \end{tabular}%
    }%
\end{table}

In the setting of motion recovery from single-view RGB video, the absence of real 3D information introduces ambiguity in the reconstructed dog mesh, \ie, the estimated results cannot reliably determine the true position and size of the dog. As shown in \cref{fig:issue}, we compare results from the same recording using single-view RGB and single-view RGB-D inputs. The reconstruction from single-view RGB shows clear deviations in body position and size compared to that using inputs with 3D cues, which aligns with the F-score results reported in \cref{tab:quan_results}. Besides, the resulting mesh from single-view RGB input also exhibits noticeable artifacts when viewed from other perspectives.

\begin{figure}[t!]
    \centering
    \begin{subfigure}{0.48\linewidth}
        \centering
        \includegraphics[width=\linewidth]{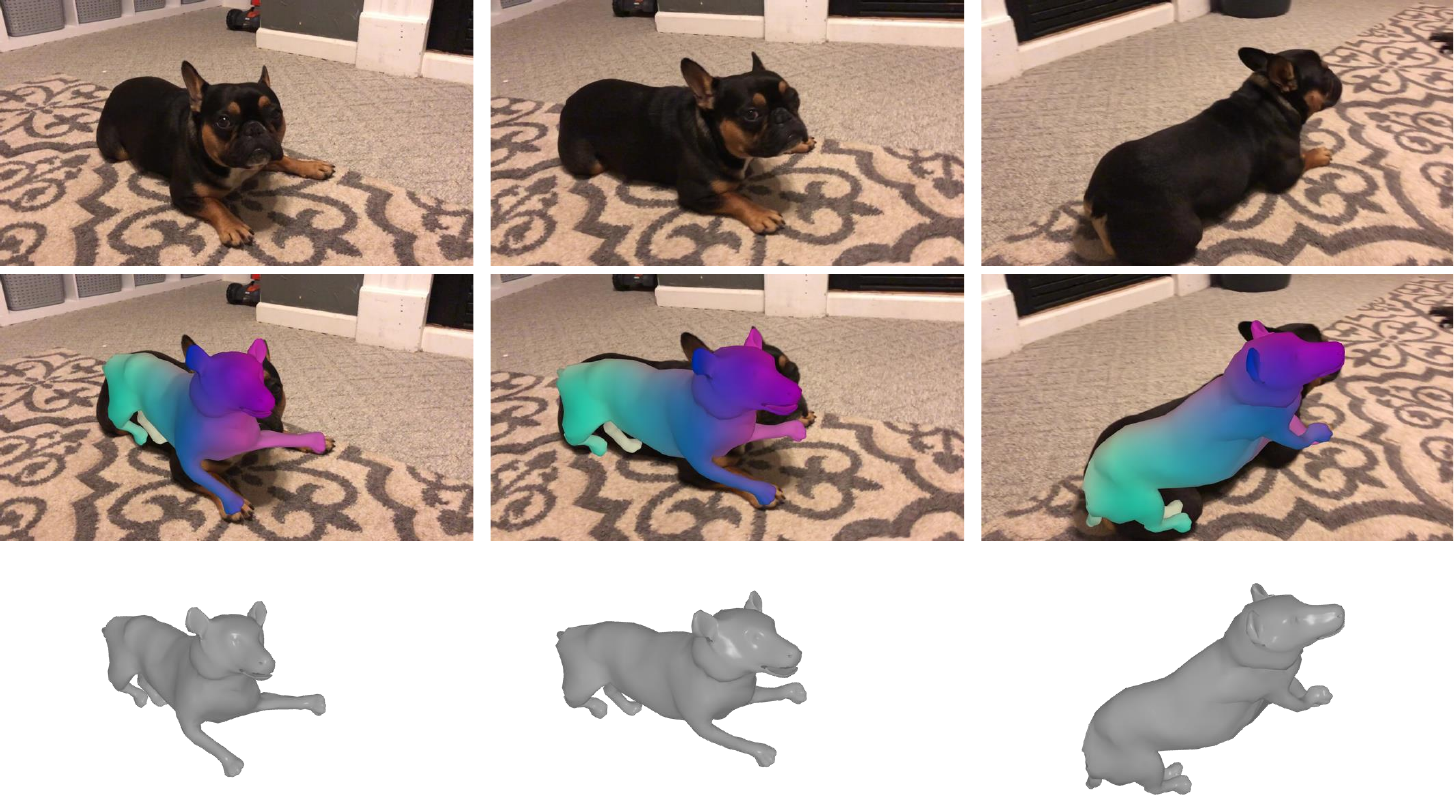}
        \caption{Results on COP3D}
        \label{fig:results_cop3d}
    \end{subfigure}\hfill%
    \begin{subfigure}{0.48\linewidth}
        \centering
        \includegraphics[width=\linewidth]{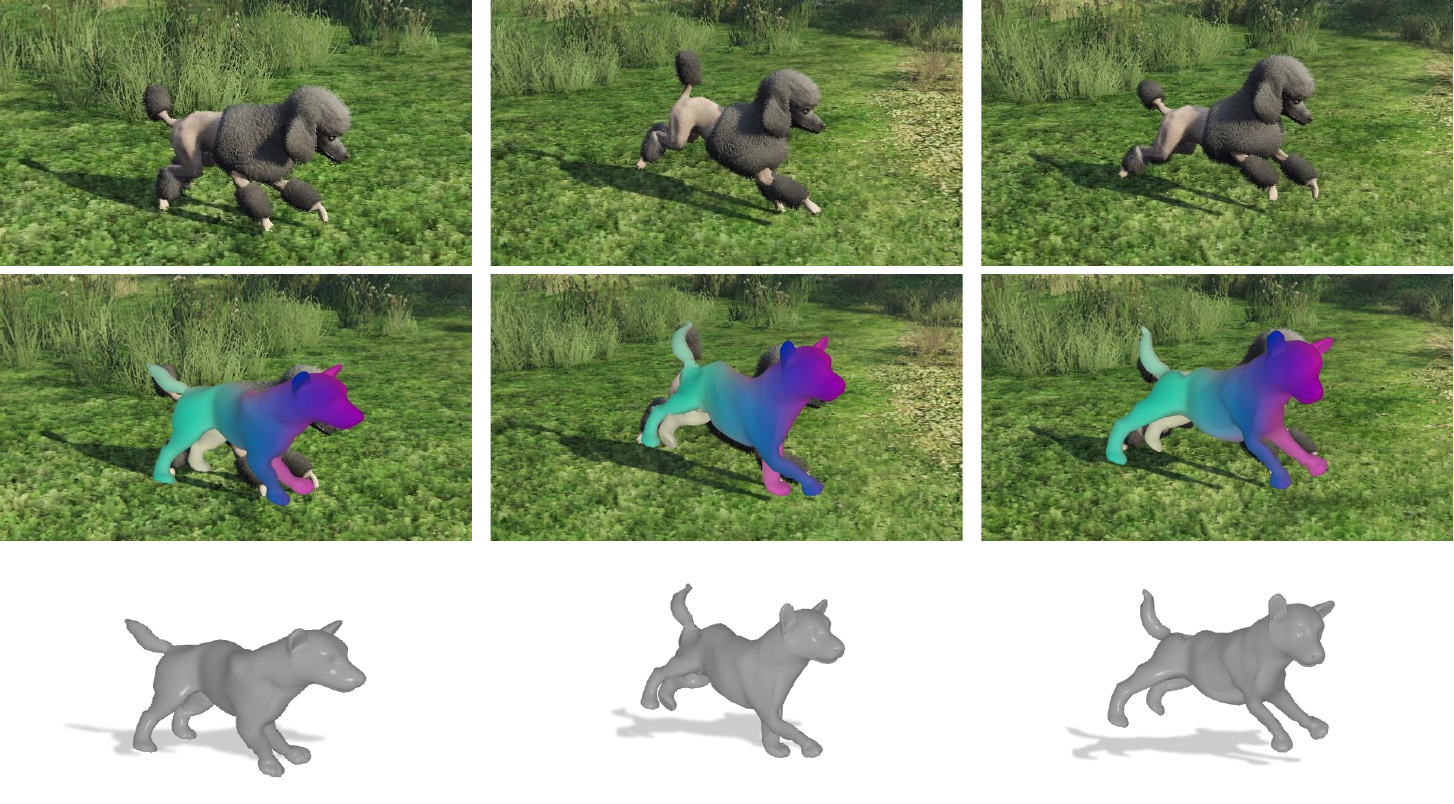}
        \caption{Results on DigiDogs}
        \label{fig:results_digidogs}
    \end{subfigure}\hfill%
    \caption{\textbf{Generalization results.} We further evaluate our method on (a) COP3D~\citep{sinha2023common} and (b) DigiDogs~\citep{shooter2024digidogs} datasets, which feature in-the-wild scenes.}
    \label{fig:generalization}
\end{figure}

\paragraph{Generalization}
We emphasize that although our method adopts an instance-specific optimization pipeline, it generalizes well across diverse cases. This generalization stems from its reliance on dog masks and keypoints predicted by pretrained models, \ie, Grounded SAM 2~\citep{ren2024grounded}, which are known for strong robustness across varied scenarios. In addition, the carefully designed cost functions are broadly applicable, as they are independent of specific pixel values and remain effective across different instances.
To further demonstrate this, we evaluate our method on the CoP3D~\citep{sinha2023common} and DigiDogs~\citep{shooter2024digidogs} datasets, which feature in-the-wild scenes. Since neither dataset provides multi-view imaging, we conduct only single-view evaluations: RGB setting on CoP3D and RGB-D setting on DigiDogs, respectively. Qualitative results are shown in \cref{fig:generalization}. For quantitative results, our method achieves IoU $0.7190$ and IoU$_{w5}$ $0.5912$ on CoP3D, and IoU $0.6644$, IoU $_{w5}$ $0.5832$, and F-score $0.7946$ on DigiDogs, which are slightly higher than those on \dataset. We attribute this improvement to two factors: (1) the dog motions in CoP3D are predominantly sprawling with limited variation, making them easier to optimize; and (2) DigiDogs is a synthetic dataset with minimal observation noise, enabling more accurate reconstruction.

\begin{table}[t!]
    \centering
    \small
    \setlength{\tabcolsep}{3pt}
    \caption{\textbf{Optimization efficiency.} Optimization time (in minutes) required by each model. Lower is better.}
    \label{tab:optim_time}
    \resizebox{\linewidth}{!}{%
        \begin{tabular}{c|ccccc|ccccc}%
        \toprule        
        Method & Setting & Stage 1 & Stage 2 & Stage 3 & Total & Setting & Stage 1 & Stage 2 & Stage 3 & Total \\
        \midrule
        AnimalAvatar & Single-view RGB & -       & -       & -       & $114.68$ & Single-view RGB-D & -       & -       & -       & $117.47$ \\
        Ours         & Single-view RGB & $2.75$  & $12.70$ & $3.33$  & $18.78$  & Single-view RGB-D & $2.83$  & $13.02$ & $3.40$  & $19.25$  \\
        \midrule
        Ours         & Multi-view RGB  & $19.82$ & $53.12$ & $11.22$ & $84.16$  & Multi-view RGB-D  & $20.38$ & $54.52$ & $11.45$ & $86.35$  \\
        \bottomrule
        \end{tabular}%
    }%
\end{table}

\paragraph{Optimization Efficiency}
To evaluate optimization efficiency, we report the runtime of our model for each stage and the entire pipeline, as shown in \cref{tab:optim_time}. Most computation is spent in the second stage, which involves all optimization terms and requires more iterations. The cost further increases with multi-view observations. Notably, our method converges in fewer iterations than AnimalAvatar~\citep{sabathier2024animal}, resulting in shorter overall runtime. This improvement stems from the strong initialization in the first stage and the use of a scaling parameter that resizes the body mesh, effectively reducing the optimization space in the second stage.

\subsection{Discussion}

\paragraph{Limitations}
While \dataset makes significant contributions, it is collected in a controlled indoor lab scenario, lacking the diversity and complexity of in-the-wild scenes. This limits the generalization of learning-based models trained solely on \dataset, particularly in scenarios with complex backgrounds, varying lighting, or occlusions.
On the algorithmic side, although our optimization pipeline is effective, it remains computationally intensive, posing challenges for real-time or large-scale deployment.

\paragraph{Future Works} In future work, we aim to scale up \dataset to support more robust and generalizable modeling of dog motion. This expansion includes augmenting the dataset with photorealistic, in-the-wild backgrounds using advanced rendering and simulation tools.
Beyond data augmentation, the expanded dataset will enable two research directions: (1) incorporating physics-based constraints to enable physically plausible motion recovery, and (2) leveraging generative models
to learn data-driven motion priors for diverse motion synthesis~\citep{yang2024omnimotiongpt,wang2025animo} and prediction tasks.

\section{Conclusion}
This work introduces \dataset, a large-scale multi-view RGB-D dataset for 4D canine motion recovery. \dataset features \ncameras cameras capturing the movements of \ndogs dogs across \nactions action types, resulting in \nmotions motion clips spanning over \nminutes minutes. Built upon this dataset, we establish four motion recovery benchmark settings to encourage future research. We also propose a unified three-stage optimization pipeline that recovers the dog's body parameters in a coarse-to-fine manner, from single-view to multi-view and from RGB to RGB-D inputs.
We believe our work lays a strong foundation for advancing understanding of animal motion, with promising applications in animation and VR/AR.

\paragraph{Acknowlwdgement}
This work is supported in part by the National Natural Science Foundation of China (NSFC) (62172043).

{
\small
\bibliographystyle{apalike}
\bibliography{reference}
}

\clearpage
\appendix
\renewcommand\thefigure{A\arabic{figure}}
\setcounter{figure}{0}
\renewcommand\thetable{A\arabic{table}}
\setcounter{table}{0}
\renewcommand\theequation{A\arabic{equation}}
\setcounter{equation}{0}

\section*{Appendix}

\section{Additional Details}

\subsection{Optimization}

\paragraph{Optimization Rate}
We set different optimization rates (\ie, learning rates) for different parameters during the optimization.
In the first stage, we initialize the optimization rates as follows: $5e^{-3}$ for the scale parameter and $5e^{-2}$ for the parameters of $\mathrm{MLP}_{TR}$. In the second stage, we set the optimization rates to $5e^{-5}$ for the scale parameter, $5e^{-2}$ for the shape parameter, and $5e^{-4}$ for both $\mathrm{MLP}_{TR}$ and $\mathrm{MLP}_{\theta}$. In the third stage, we set the optimization rates to $1e^{-5}$ for both $\mathrm{MLP}_{TR}$ and $\mathrm{MLP}_{\theta}$. All optimization rates follow a step-wise decay schedule: at $75\%$ and $93.75\%$ of the total optimization steps, the rate is multiplied by a factor of $0.3$.

\paragraph{Loss Weight} 
For each optimization loss term, we set the loss weight as follows: $400.0$ for $\mathcal{L}_t^{\mathrm{mask}}$, 
$60.0$ for $\mathcal{L}^{\mathrm{kp}}_t$, 
$1.0$ for $\mathcal{L}_t^{\mathrm{depth}}$, 
$20.0$ for $\mathcal{L}^{\mathrm{cse}}_t$, 
$2.5$ for $\mathcal{L}_{t}^{\mathrm{cross}}$, 
$0.005$ for $\mathcal{L}_t^{\mathrm{prior}}$, 
and $0.1$ for $\mathcal{L}^{\mathrm{temp}}$.

\subsection{Shape \& Pose Prior Loss}

To enhance the plausibility of the estimated results and mitigate implausible joint poses or abnormal body proportions, we incorporate two prior constraint losses following existing studies~\citep{ruegg2022barc, zuffi20173d}: a pose prior loss and a shape prior loss. These priors are modeled as Gaussian distributions derived from the statistical properties of animal datasets. The Mahalanobis distance is employed during optimization to regularize the parameters, guiding the SMAL predictions toward anatomically consistent pose and body shape distributions.

\paragraph{Shape Prior Loss} The shape prior loss $\mathcal{L}^{\text{shape}}$ consists of two components: a body shape prior loss $\mathcal{L}^{\text{body}}$ and a limb proportion prior loss $\mathcal{L}^{\text{limb}}$.
The body shape prior loss $\mathcal{L}^{\text{body}}$ is derived from a Gaussian prior fitted to the body shape parameters in the BARC~\citep{ruegg2022barc} dataset. Here, $\bm{\mu}_{\text{body}}$ and $\bm{\Sigma}_{\text{body}}$ represent the mean vector and the positive semi-definite covariance matrix computed from the data.
The Mahalanobis distance between the optimized shape vector $\bm{\beta}$ and the prior distribution is then given by
\begin{equation}
    \mathcal{L}^{\text{body}} =  (\bm{\beta} - \bm{\mu}_{\text{body}})^\top \bm{\Sigma}_{\text{body}}^{-1} (\bm{\beta} - \bm{\mu}_{\text{body}}).    
\end{equation}
To further ensure reasonable limb proportions, we introduce a limb proportion prior loss $\mathcal{L}^{\text{limb}}$, which applies a weighted quadratic penalty to limb-related shape coefficients, \ie,
\begin{equation}
    \mathcal{L}^{\text{limb}} = \left\| \mathbf{w} \odot \bm{\beta}^{\text{limb}} \right\|_2^2,
\end{equation}
where $\mathbf{w}$ is a manually defined weight vector.
The final shape prior loss combines the body and limb prior terms through a weighted sum, \ie, $\mathcal{L}^{\text{shape}} = w_\text{body} \cdot \mathcal{L}^{\text{body}} + w_{\text{limb}} \cdot \mathcal{L}^{\text{limb}}$.

\paragraph{Pose Prior Loss} Similarly, we model the joint rotation prior as a Gaussian distribution characterized by the mean vector $\bm{\mu}_{\text{pose}}$ and a positive semi-definite covariance matrix $\bm{\Sigma}_{\text{pose}}$.
The Mahalanobis distance between the optimized pose vector and this prior distribution is then computed as
\begin{equation}
    \mathcal{L}_{t}^{\text{pose}} = (\theta_t - \bm{\mu}_{\text{pose}})^\top \bm{\Sigma}_{\text{pose}}^{-1} (\theta_t - \bm{\mu}_{\text{pose}}).
\end{equation}

\subsection{Metric Computation}

\paragraph{F-score}
Following \citet{wang2018pixel2mesh}, we first uniformly sample points from our optimization result and lift the masked depth image pixels into 3D space.
We calculate precision and recall by measuring the percentage of points from the optimization or the depth image reprojection that have a nearest neighbor in the other set within a specified threshold. We set the threshold as $0.05$. The F-score is then computed as the harmonic mean of precision and recall.

\paragraph{Pene$_\%$} We compute the penetration ratio as the proportion of SMAL mesh vertices that penetrate the ground plane in each frame.

\paragraph{Jitter} The jitter metric quantifies the temporal smoothness of predicted SMAL joint trajectories, indicating the stability of the motion. It is calculated as the mean second-order difference of all joint positions across the sequence.

\paragraph{Foot Skating} Foot skating measures the unintended horizontal displacement of the feet while in contact with the ground, indicating unrealistic gait motion. The metric is calculated by averaging the horizontal velocity of the feet whenever their distance to the ground plane is below $0.04\mathrm{m}$.

\section{Dataset Statistics}

\begin{figure}[t!]
    \centering
    \begin{subfigure}{\linewidth}
        \centering
        \includegraphics[width=0.8\linewidth]{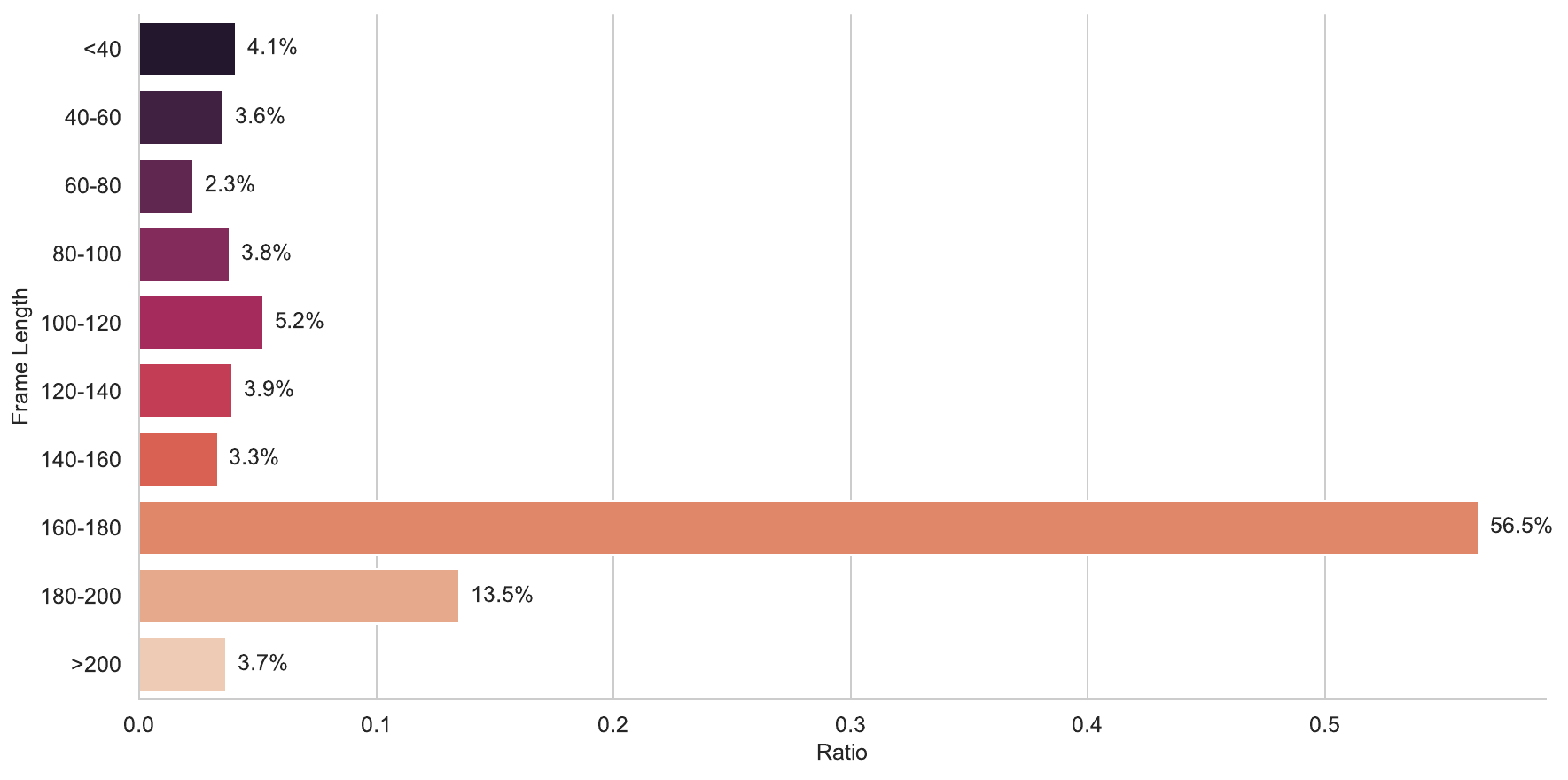}
        \caption{Frame Length Distribution}
        \label{supp:fig:frame}
    \end{subfigure}\hfill%
    \begin{subfigure}{\linewidth}
        \centering
        \includegraphics[width=0.8\linewidth]{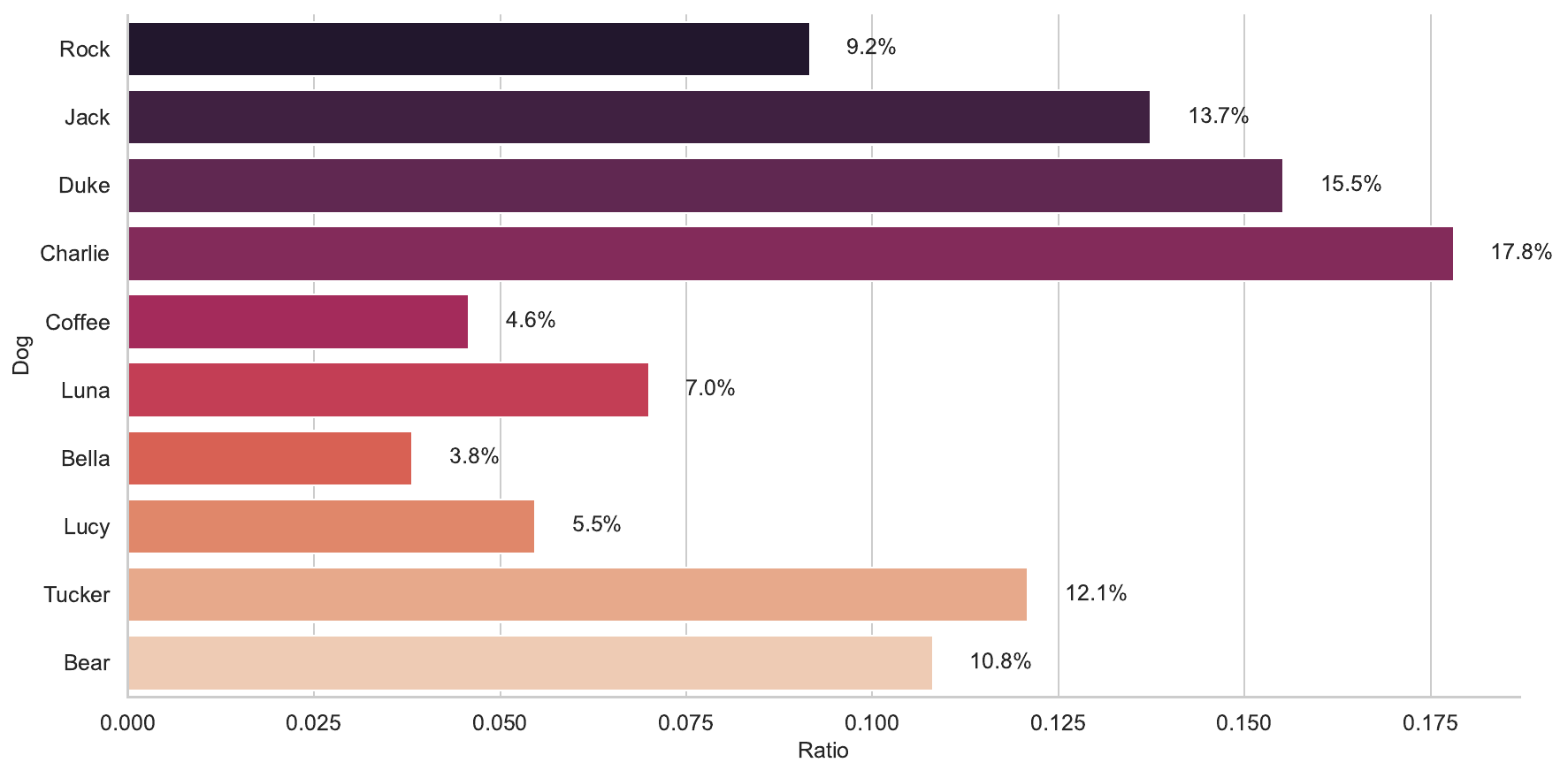}
        \caption{Dog Instance Frequency}
        \label{supp:fig:dog}
    \end{subfigure}\hfill%
    \begin{subfigure}{\linewidth}
        \centering
        \includegraphics[width=0.8\linewidth]{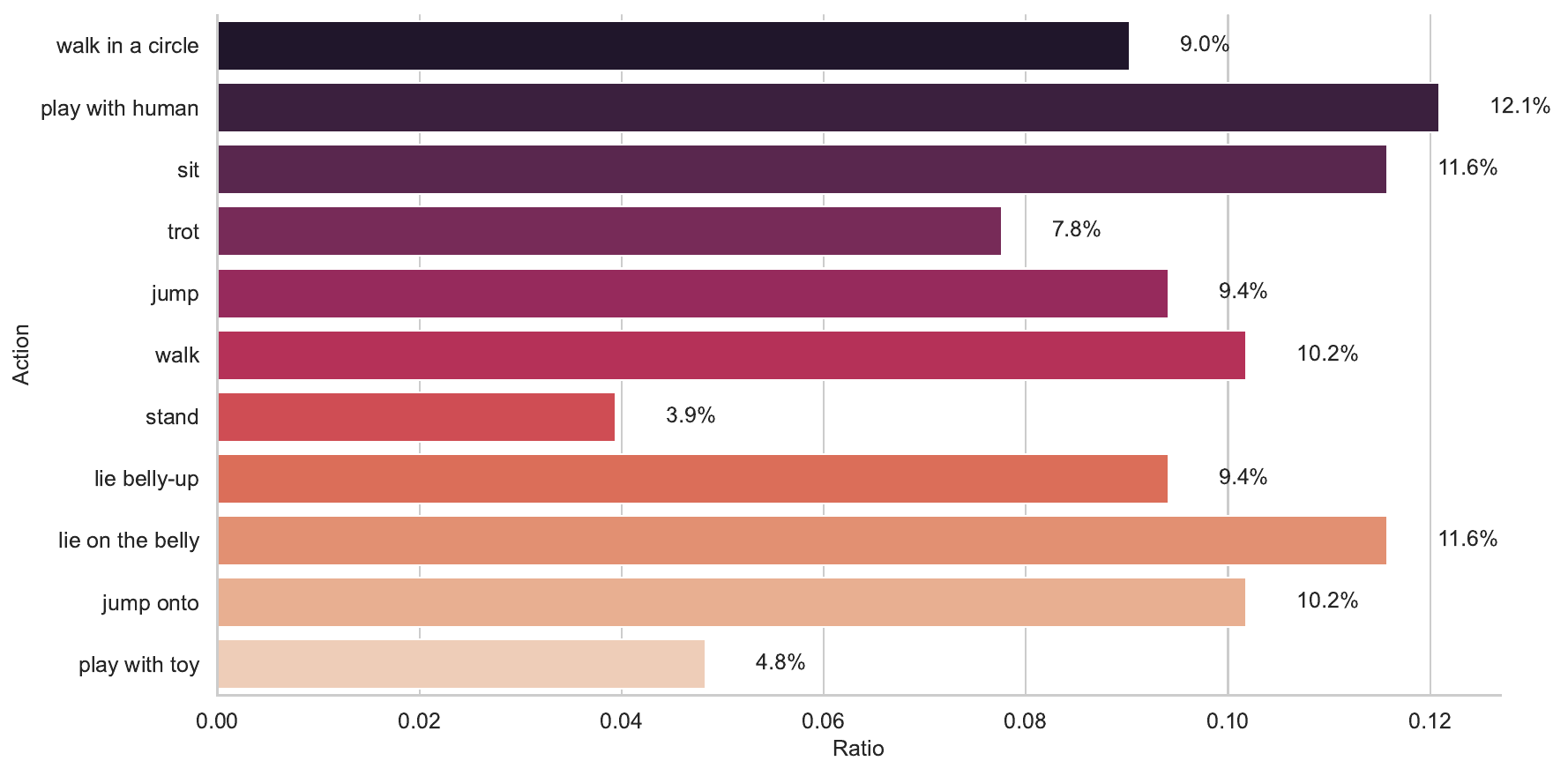}
        \caption{Action Frequency}
        \label{supp:fig:action}
    \end{subfigure}\hfill%
    \caption{Statistics of \dataset. Notably, the action ``stand'' denotes the dog standing upright on its hind legs.}
    \label{supp:fig:statistics}
\end{figure}

\paragraph{Statistics}
We provide additional statistics for the \dataset dataset in \cref{supp:fig:statistics}.
\Cref{supp:fig:frame} illustrates the distribution of motion sequence lengths. In \dataset, frame counts range from $13$ to $286$, with over half of the sequences falling between 180 and 200 frames. The average frame length is approximately $155$.
\Cref{supp:fig:dog} presents the distribution of motion data across \ndogs individual dog instances in \dataset. Some dogs, \eg, Charlie, appear more responsive to the trainers, so we collect more motion sequences from those instances.
Additionally, each motion sequence in \dataset is annotated with one of \nactions predefined action categories. The motion distribution over these action types is shown in \cref{supp:fig:action}.

\begin{figure}[t!]
    \centering
    \includegraphics[width=\linewidth]{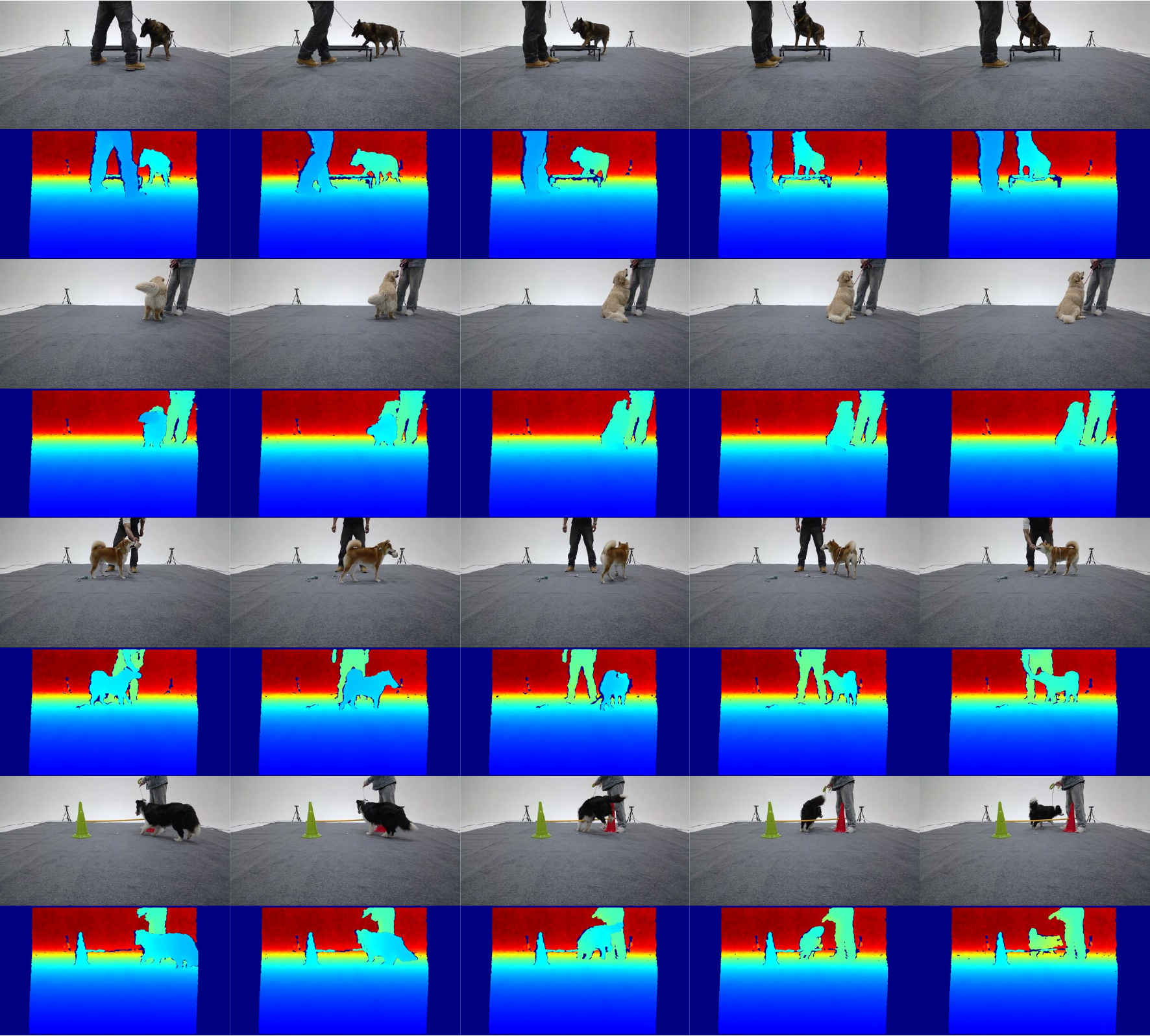}
    \caption{\textbf{Motion recording visualization.} We visualize four motion recordings, each presenting both an RGB sequence and a depth sequence.}
    \label{supp:fig:motion}
\end{figure}

\paragraph{Visualization of Motion Recordings} We additionally provide visualizations of the motion recordings in \cref{supp:fig:motion}. For video demonstrations, please refer to the supplementary demo video.

\section{Additional Results}

\paragraph{Qualitative Result Comparison}

\begin{figure}[ht!]
    \centering
    \begin{subfigure}{\linewidth}
        \centering
        \includegraphics[width=0.25\linewidth]{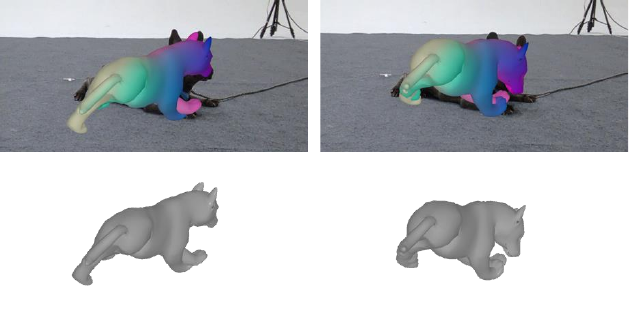}
        \caption{Optimization Results of SMALify}
        \label{fig:smalify}
    \end{subfigure}\hfill%
    \vspace{6pt}
    \begin{subfigure}{\linewidth}
        \centering
        \includegraphics[width=0.5\linewidth]{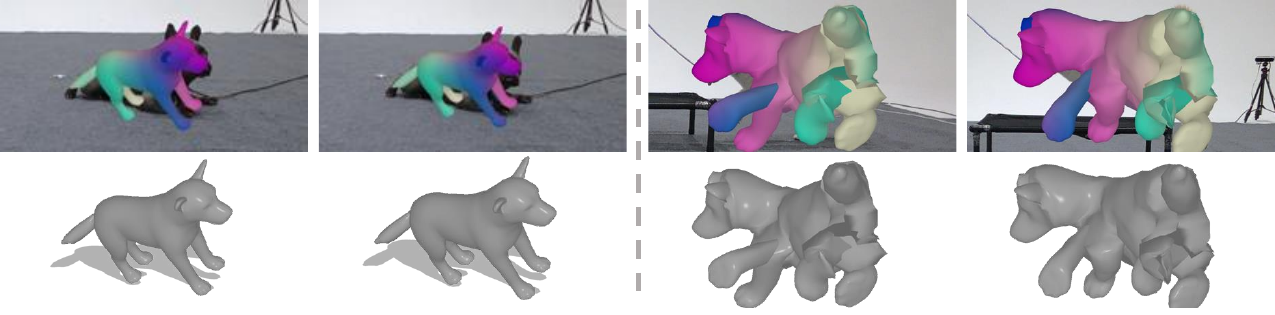}
        \caption{Optimization Results of AnimalAvatar}
        \label{fig:animal_avatar}
    \end{subfigure}\hfill%
    \vspace{6pt}
    \begin{subfigure}{\linewidth}
        \centering
        \includegraphics[width=\linewidth]{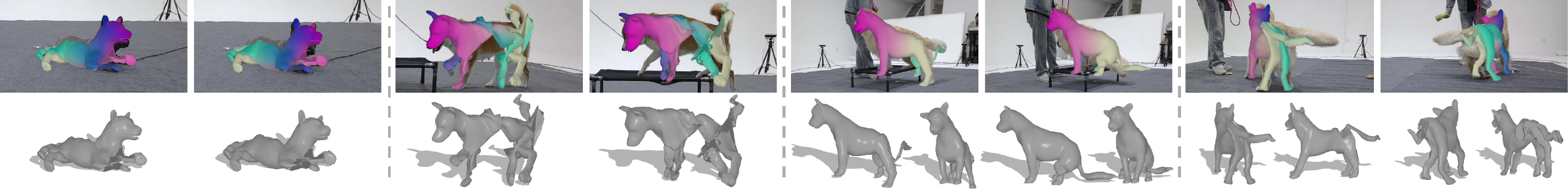}
        \caption{Optimization Results of \textit{w/o stage 1}}
        \label{fig:wo_stage1_result}
    \end{subfigure}\hfill%
    \vspace{6pt}
    \begin{subfigure}{\linewidth}
        \centering
        \includegraphics[width=\linewidth]{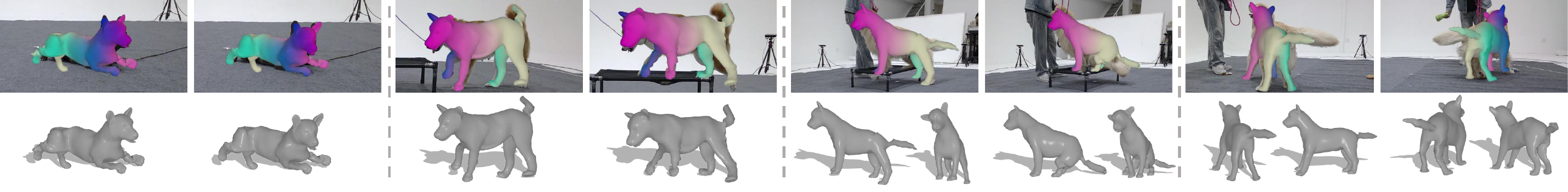}
        \caption{Optimization Results of \textit{w/o stage 3}}
        \label{fig:wo_stage3_result}
    \end{subfigure}\hfill%
    \vspace{6pt}
    \begin{subfigure}{\linewidth}
        \centering
        \includegraphics[width=\linewidth]{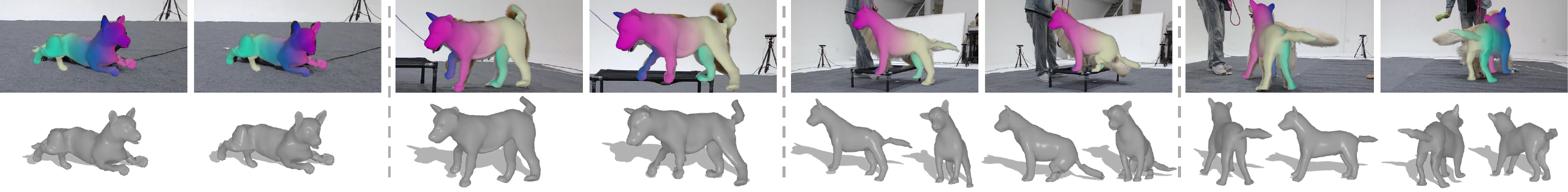}
        \caption{Optimization Results of Our Full Method}
        \label{fig:full_method_result}
    \end{subfigure}\hfill%
    \caption{\textbf{Qualitative result comparison.} (a) For the optimization results of SMALify, we present motion recovery from single-view RGB images. (b) For the optimization results of AnimalAvatar, we present motion recovery from single-view RGB images and single-view RGB-D images. (c) (d) For the other baseline methods, we show results under four settings, \ie, single-view RGB, single-view RGB-D, multi-view RGB, and multi-view RGB-D, arranged from left to right. For each setting, we show two frames from the motion sequence.}
    \label{supp:fig:qual_result_comparison}
\end{figure}

In \cref{supp:fig:qual_result_comparison}, we compare the qualitative results of the baseline models with those of our full model. When recovering dog motion from a single-view RGB image, the comparison shows that both SMALify~\citep{biggs2020left} and AnimalAvatar~\citep{sabathier2024animal} fail to capture the dynamics of body parts, particularly the leg movements, resulting primarily in global translation and orientation changes. Without the initialization provided by the first stage, the method struggles to reconstruct the complete motion.
We also provide video demonstrations in the supplementary demo video.

\begin{figure}[ht!]
    \centering
    \includegraphics[width=0.5\linewidth]{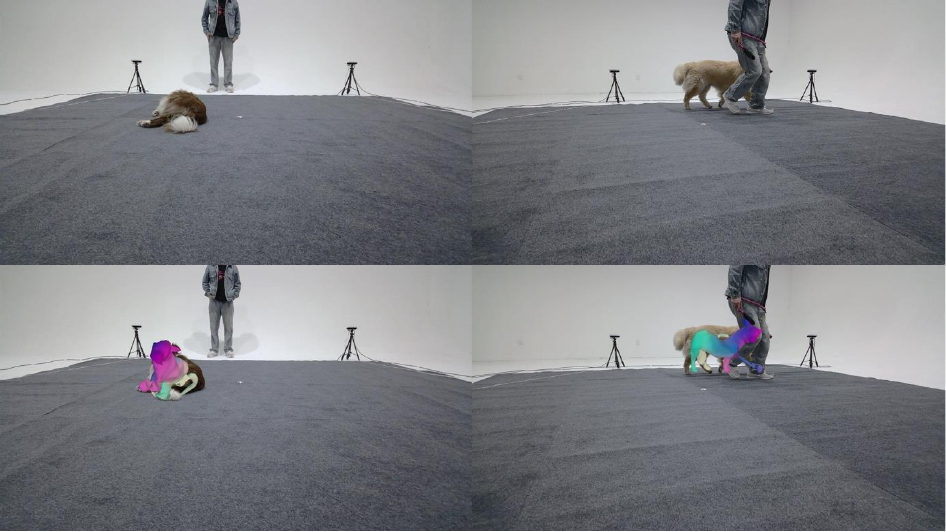}
    \caption{\textbf{Failure cases.}}
    \label{supp:fig:failure_cases}
\end{figure}

\paragraph{Failure Cases}
We present several failure cases produced by our method in \cref{supp:fig:failure_cases}. As shown in the results, when the dog is curled up in a resting position, it becomes difficult to distinguish individual body parts due to the uniform texture across the body, leading to inaccurate reconstruction.
Additionally, when the dog is occluded by the dog trainer, particularly the distinguishable body parts such as the head, the limited visual information also leads to poor reconstruction of the body shape.

\section{Additional Experiments}

\paragraph{Impact of Different Initial $s$ in \textit{w/o stage 1}} As shown in \cref{tab:initial_scale}, the model \textit{w/o stage 1} exhibits large variations in Pene$_\%$ and Foot Skating under different scaling factors $s$. When $s=0.3$, the model tends to retain a smaller body size, causing the mesh to float above the floor. This leads to reduced penetration and fewer contacts, and non-contact meshes inherently yield zero penetration and zero foot skating. Conversely, when $s=1.0$, the resulting meshes penetrate deeply into the floor, producing a higher penetration ratio. Excessive penetration also reduces the number of valid contact frames for foot skating computation, thereby lowering the averaged foot skating score.
Both variants show low Jitter. We speculate that without stage 1 optimization, the pipeline primarily adjusts global translation and orientation in the second stage while neglecting high-frequency motion details. As a result, most joints move more smoothly with reduced dynamics, leading to lower jitter values.

\paragraph{Simpler Temporal Smoothing Alternatives}
We replace the smoothness optimization with simpler Kalman and spline smoothing in the third stage, which yields smoother results with reduced jitter. However, they also result in slightly lower IoU and F-score. The results are presented in \cref{supp:tab:smooth}.

\begin{table}[t!]
    \centering
    \small
    \setlength{\tabcolsep}{3pt}
    \caption{\textbf{Impact of different initial $s$ in \textit{w/o stage 1}.}}
    \label{tab:initial_scale}
        \begin{tabular}{c|ccc}%
            \toprule
            \multirow{2.4}{*}{Method} & \multicolumn{3}{c}{Single-view RGB} \\
            \cmidrule(lr){2-4} & Pene$_\%$ $\downarrow$ & Jitter $\downarrow$ & FS $\downarrow$ \\
            \midrule
            w/o stage 1 ($s=0.3$)  & $0.0796$ & $0.0056$ & $0.0019$ \\
            w/o stage 1 ($s=1.0$)  & $0.4077$ & $0.0096$ & $0.0147$ \\
            \bottomrule
        \end{tabular}%
\end{table}

\begin{table}[t!]
    \centering
    \caption{\textbf{Quantitative results with simpler temporal smoothing alternatives.}}
    \label{supp:tab:smooth}
        \begin{tabular}{cccccc}%
            \toprule
            Setting & Method & IoU $\uparrow$ & $\text{IoU}_{\text{w5}}$ $\uparrow$ & F-score $\uparrow$ & Jitter $\downarrow$ \\
            \midrule
            Single-view RGB   & Kalma  & $0.6251$ & $0.5015$ & $0.3119$ & $0.0211$ \\
                              & Spline & $0.4821$ & $0.2968$ & $0.3426$ & $0.0047$ \\
                              & Ours   & $0.6505$ & $0.5247$ & $0.3593$ & $0.0249$ \\
            \midrule
            Single-view RGB-D & Kalman & $0.6248$ & $0.4945$ & $0.6889$ & $0.0178$ \\
                              & Spline & $0.4812$ & $0.3031$ & $0.6951$ & $0.0037$ \\
                              & Ours   & $0.6414$ & $0.5228$ & $0.7673$ & $0.0162$ \\
            \midrule
            Multi-view RGB    & Kalman & $0.6021$ & $0.3937$ & $0.8586$ & $0.0165$ \\
                              & Spline & $0.4769$ & $0.2379$ & $0.8080$ & $0.0036$ \\
                              & Ours   & $0.6258$ & $0.4490$ & $0.9107$ & $0.0142$ \\
            \midrule
            Multi-view RGB-D  & Kalman & $0.5907$ & $0.3847$ & $0.8595$ & $0.0159$ \\
                              & Spline & $0.4748$ & $0.2337$ & $0.8089$ & $0.0045$ \\
                              & Ours   & $0.6286$ & $0.4549$ & $0.9137$ & $0.0143$ \\
            \bottomrule
        \end{tabular}%
\end{table}

\section{IRB (Institutional Review Board)} 

This study does not involve human subjects or physiological experiments. The dataset in this work is collected from videos of dogs in natural environments without invasive procedures or interventions. According to our institution’s research ethics policy, IRB (Institutional Review Board) approval is only required for studies involving human participants or biological experimentation. Therefore, no IRB approval is necessary or applicable for this research.

\section{Broader Impact}

This work focuses on reconstructing the motion of dogs from visual data, contributing to the broader understanding of animal motion in computer vision and machine learning. Potential positive societal impacts include enabling improved animal behavior analysis and advancing animation or simulation technologies for education, film, and virtual environments. Additionally, the data and techniques may support future research in biologically inspired locomotion in robotics. Given the nature of the work, we do not anticipate any significant negative societal impacts.

\end{document}